\documentclass[acmtog, nonacm]{acmart}
\acmSubmissionID{858}

\usepackage{booktabs} 

\usepackage{xcolor}
\usepackage{colortbl}

\usepackage[ruled]{algorithm2e} 

\SetAlFnt{\small}
\SetAlCapFnt{\small}
\SetAlCapNameFnt{\small}
\SetAlCapHSkip{0pt}

\acmJournal{TOG}

\copyrightyear{2025}
\acmYear{2025}
\setcopyright{cc}
\setcctype{by}
\acmConference[SIGGRAPH Conference Papers '25]{Special Interest Group on Computer Graphics and Interactive Techniques Conference Conference Papers }{August 10--14, 2025}{Vancouver, BC, Canada}
\acmBooktitle{Special Interest Group on Computer Graphics and Interactive Techniques Conference Conference Papers (SIGGRAPH Conference Papers '25), August 10--14, 2025, Vancouver, BC, Canada}
\acmDOI{10.1145/3721238.3730710}
\acmISBN{979-8-4007-1540-2/2025/08}

\usepackage{xfrac}

\usepackage{float}
\usepackage{soul}
\usepackage{dblfloatfix}

\definecolor{lightred}{RGB}{237.,146.,191.}

\definecolor{lightorange}{RGB}{255.0,176.0,127.0 }

\definecolor{lightyellow}{RGB}{255.0,215.0,127.0}

\DeclareRobustCommand{\hlredcaption}[1]{{\sethlcolor{lightred}\hl{#1}}}
\DeclareRobustCommand{\hlorangecaption}[1]{{\sethlcolor{lightorange}\hl{#1}}}
\DeclareRobustCommand{\hlyellowcaption}[1]{{\sethlcolor{lightyellow}\hl{#1}}}

\newcommand{\hlred}[1]{\cellcolor{lightred}{#1}}
\newcommand{\hlorange}[1]{\cellcolor{lightorange}{#1}}
\newcommand{\hlyellow}[1]{\cellcolor{lightyellow}{#1}}

\newcommand{\rev}[1]{#1}

\newcommand{\eg}{\emph{e.g.}}

\newcommand{\etal}{\emph{et~al.}}

\newcommand{\reffig}[1]{Fig.~\ref{#1}}

\newcommand{\refeq}[1]{Eq.~\ref{#1}}
\newcommand{\refsec}[1]{Sec.~\ref{#1}}

\newcommand{\hiddencomment}[1]{\iftrue {\color{gray}#1}\else {}\fi}  
\newcommand{\suppcomment}[1]{\iftrue {\color{red}[move to supp:] #1}\else {}\fi}  
\newcommand{\editcomment}[1]{\iftrue {\color{green}[edit:] #1}\else {}\fi}

\newcommand{\patchwidth}{0.28\columnwidth}

\begin{document}
\title[TeGA]{TeGA: Texture Space Gaussian Avatars for High-Resolution Dynamic Head Modeling}


\author{Gengyan Li}
\email{gengyan.li@inf.ethz.ch}
\orcid{0000-0002-1427-7612}
\affiliation{%
  \institution{ETH Zurich}
  \city{Zurich}  
  \country{Switzerland}}
\affiliation{%
  \institution{Google}
  \city{Zurich}
  \country{Switzerland}}

\author{Paulo Gotardo}
\email{gotardo@google.com}
\orcid{0000-0001-8217-5848}
\affiliation{%
  \institution{Google}
  \city{Zurich}
  \country{Switzerland}}

\author{Timo Bolkart}
\email{tbolkart@google.com}
\orcid{0000-0002-3829-3924}
\affiliation{%
  \institution{Google}
  \city{Zurich}
  \country{Switzerland}}

\author{Stephan Garbin}
\email{stephangarbin@google.com}
\orcid{0009-0000-5005-8110}
\affiliation{%
  \institution{Google}
  \city{London}
  \country{United Kingdom}}

\author{Kripasindhu Sarkar}
\email{krsarkar@google.com}
\orcid{0000-0002-0220-0853}
\affiliation{%
  \institution{Google}
  \city{Zurich}
  \country{Switzerland}}

\author{Abhimitra Meka}
\email{abhim@google.com}
\orcid{0000-0001-7906-4004}
\affiliation{%
  \institution{Google}
  \city{San Francisco}
  \country{United States}}

\author{Alexandros Lattas}
\orcid{0000-0002-9964-6105}
\email{lattas@google.com}
\affiliation{%
  \institution{Google}
  \city{Zurich}
  \country{Switzerland}}

\author{Thabo Beeler}
\orcid{0000-0002-8077-1205}
\email{tbeeler@google.com}
\affiliation{%
  \institution{Google}
  \city{Zurich}
  \country{Switzerland}}

\renewcommand{\shortauthors}{Li, G. et al.}

\begin{abstract}
   Sparse volumetric reconstruction and rendering via 3D Gaussian splatting have recently enabled animatable 3D head avatars that are rendered under arbitrary viewpoints with impressive photorealism. Today, such photoreal avatars are seen as a key component in emerging applications in telepresence, extended reality, and entertainment. Building a photoreal avatar requires estimating the complex non-rigid motion of different facial components as seen in input video images; due to inaccurate motion estimation, animatable models typically present a loss of fidelity and detail when compared to their non-animatable counterparts, built from an individual facial expression. Also, recent state-of-the-art models are often affected by memory limitations that reduce the number of 3D Gaussians used for modeling, leading to lower detail and quality. To address these problems, we present a new high-detail 3D head avatar model that improves upon the state of the art, largely increasing the number of 3D Gaussians and modeling quality for rendering at 4K resolution. Our high-quality model is reconstructed from multiview input video and builds on top of a mesh-based 3D morphable model, which provides a coarse deformation layer for the head. Photoreal appearance is modelled by 3D Gaussians embedded within the continuous UVD tangent space of this mesh, allowing for more effective densification where most needed. Additionally, these Gaussians are warped by a novel UVD deformation field to capture subtle, localized motion. Our key contribution is the novel deformable Gaussian encoding and overall fitting procedure that allows our head model to preserve appearance detail, while capturing facial motion and other transient high-frequency features such as skin wrinkling.
\end{abstract}

%
%

%
%


\begin{teaserfigure}
	\centering
	\includegraphics[width=\textwidth]{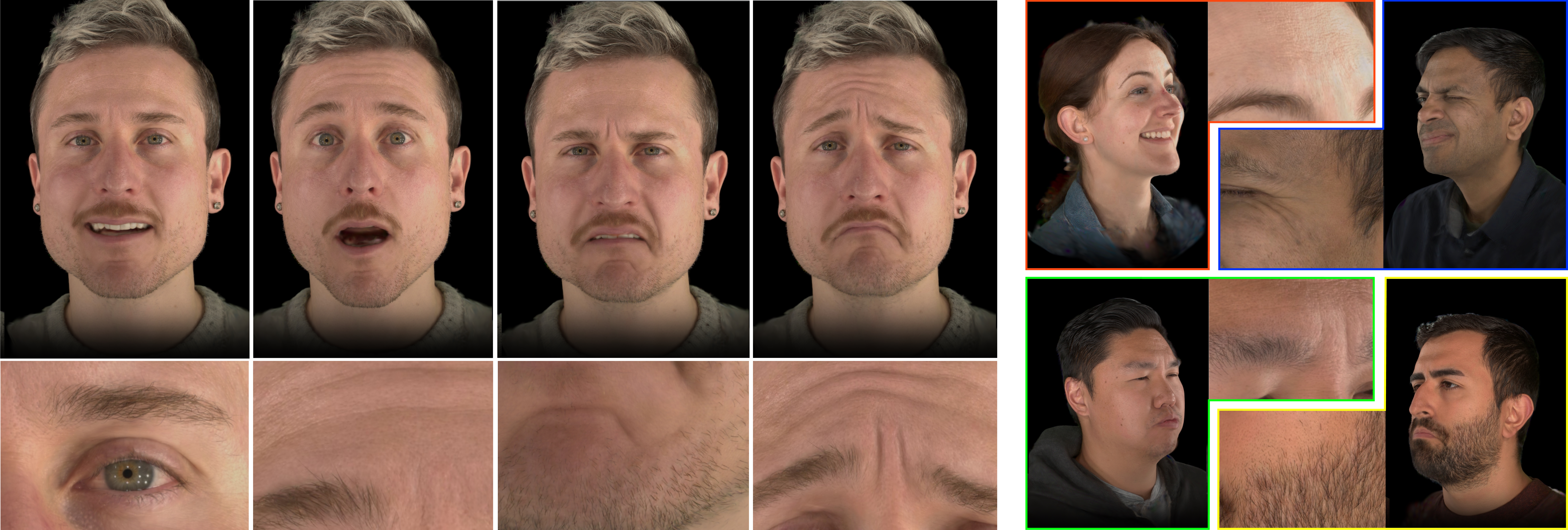}
	\caption{We propose a novel method that builds photorealistic 3D head avatars for rendering at high resolution with controllable expressions and viewpoints. 
		Given a deformable 3D mesh model, our method effective learns a sparse, expression-dependent volumetric texture for rendering via 3D Gaussian splatting. It effectively leverages densification where most needed, estimating up to 4 million 3D Gaussians within the continuous UVD tangent space of the mesh. In this figure, we show multiple high fidelity novel face expression reconstructions.}
	\Description{This is the teaser figure for the article.}
	\label{fig:teaser}
\end{teaserfigure}

\maketitle

\section{Introduction}

The photorealistic modeling of human faces 
has enjoyed widespread attention in graphics and computer vision communities,
and its applications have been impactful in numerous industries,
including telecommunications, media, gaming and medicine to name a few.
Telepresence, once science fiction, appears closer and closer,
as hardware and algorithms are making great strides.
In the pursuit to bridge the uncanny valley,
numerous recent works have achieved pleasing high-resolution human animation.
However, when considering very high resolution rendering as a requirement, for example, in visual effects in contemporary movies, many challenges arise in the accurate reconstruction of details,
which are paramount to achieve true likeness for a digital human.

A seminal step towards head modeling was the introduction of 3D Morphable Models (3DMM) \cite{blanz1999}, which are still used over twenty years later.
Controllable illumination capturing rigs \cite{debevec2012light} enabled the acquisition
of high-quality and relightable facial appearance, though generation and manipulation has remained a challenge.
With the introduction of Generative Adversarial Networks (GAN) \cite{goodfellow2014generative},
and especially StyleGAN \cite{karras2020analyzing},
the research community witnessed a big leap towards high-quality facial appearance modeling
in camera-space \cite{nagano2018pagan}, UV-space \cite{gecer2019ganfit}, or in 3D \cite{chan2022efficient}.
In addition, Latent Diffusion Models \cite{rombach2022high} significantly improved such workstreams. 
At representation level,
Neural Radiance Fields (NeRFs) \cite{mildenhall2021nerf} revolutionized the space, providing a flexible medium
for holistic human reconstruction and animation \cite{park2021nerfies, Zielonka2022Insta, hong2021headnerf, sarkar2023litnerf},
but were restricted by slow speeds and their difficulty to be manipulated. 
3D Gaussian Splatting (3DGS) \cite{kerbl20233d} recently filled these gaps,
with an explicit representation with orders of magnitude faster optimization and unprecedented visual quality.

Multiple works have since explored with increasing success the space of high-quality 3DGS head modeling, following mostly two different paradigms:
a) 3DMM-attached Gaussians (e.g.~\cite{Qian2024gaussianavatars, giebenhain2024npga}), and
b) CNN-based generated Gaussians (e.g.~\cite{saito2024relightable, teotia2024gaussianheads}).
Both approaches still underperform in producing sharp and detailed images when closely inspected. As such, an adequate exploration into the domain of very high resolution rendering is still needed.
On one hand, 3DMM-based approaches are limited by the underlying linear model, or require the addition of expensive dense MLPs that learn to deform the Gaussians, and typically remain underconstrained and underperform in challenging cases.
On the other hand, CNN-based methods generate a set number of Gaussians on a UV map of limited resolution, and cannot also adaptively allocate details, \eg,~for facial hair.
As we show in our extensive experiments,
both cases perform poorly when attempting to capture fine facial details
in high resolution.

Our work proposes a new hybrid approach for rendering digital head avatars at very high resolution (4K). 
To capture and animate fine facial details, our method effectively combines a 3DMM mesh, adaptive 3D Gaussian densification, and an expression-dependent deformation model that is agnostic to the number of 3D Gaussians.
We achieve high-detail, 4K renderings without requiring super-resolution networks that operate on the camera image plane.
Even though this is claimed by works similar to ours \cite{Qian2024gaussianavatars, saito2024relightable}, 
our extensive qualitative and quantitative experiments showcase the importance of our improvements,
especially around the faithful depiction of facial details.
In summary:
\begin{enumerate}
\item We propose a simple approach for rigging 3D Gaussians within the continuous tangent space of 3DMM face models, allowing Gaussians to move freely across mesh triangles.
\item We propose a novel CNN-based deformation model that is agnostic to the number of 3D Gaussians, naturally enabling adaptively densification of the representation to improve detail where most needed, with expression-dependent shading.
\item We show significant improvements over baseline SOTA methods, and demonstrate the ability to render even extreme closeup images at high quality.
\end{enumerate}
\section{Related Work}
\label{sec:related_work}

\subsection{Head Modeling}
\label{sec:related_head_modeling}

Head modeling has been a very active field of research, arguably materialising with the seminal 3D Morphable Models work \cite{blanz1999, egger20203d}, a PCA-based model of 3D face shape and appearance.
Numerous works have followed, extending the quality and generalization of 3DMMs \cite{paysan20093d, booth20163d, li2017flame}.
Appearance has been captured with linear models \cite{smith2020morphable},
GANs \cite{gecer2019ganfit, lattas2020avatarme, lattas2023fitme, li2020learning, luo2021normalized},
and diffusion models \cite{zhang2023dreamface, papantoniou2023relightify}, with increasing quality.
Nevertheless, non-skin areas pose a challenge for such models.

NeRFace~\cite{Gafni2021nerface} combined a 3DMM with Neural Radiance Fields \cite{mildenhall2021nerf}, which provided a more flexible representation capable of also modeling non-skin regions, achieved higher quality appearance modeling.
Several follow up works extend this to using a layered NeRF representation \cite{shellnerf2024}, create fast trainable avatars  \cite{Zielonka2022Insta}, or build NeRF-based parametric head models \cite{hong2021headnerf,zhuang2022mofanerf}.
Despite their qualitative success, NeRF-based methods often lack high-frequency details, and suffer from slow evaluation.
Our method instead uses 3D Gaussian Splatting to represent the neural, leading in higher quality and rendering performance.

\subsection{3DGS-based Head Modeling}
\label{sec:related_3dgs}

The recent work of 3D Gaussian Splatting \cite{kerbl20233d} enabled radiance field training and rendering at much faster rates, and was shown to accurately track and reconstruct human motion \cite{luiten2024dynamic}.
A main paradigm for 3DGS head modeling, introduced by GaussianAvatars \cite{Qian2024gaussianavatars}, is to rig the 3D Gaussians on the triangles of 3DMMs, which can be optimized and densified locally, while being animated by 3DMM parameters.
NPGA \cite{giebenhain2024npga} further extended this approach with non-linear 3DMMs, adaptive density for occluded facial areas, and learned latent features that enhance the learning of complex facial dynamic behaviour.
The attachment of the Gaussians to the mesh may also be learned for additional flexibility \cite{shao2024splattingavatar, zhao2024psavatar}.
Moreover, Rig3DGS \cite{rivero2024rig3dgs} introduced a prior-based learnable deformation, to achieve the synthesis of scene and the head, while trained only on casually captured videos.
3DMM-based representations have also been extended with MLPs in order to model expression depended effects such as wrinkling \cite{Dhamo2024headgas, xu2023gaussianheadavatar}.
Finally, hybrid approaches have also been proposed \cite{xiao2024bridging}, to leverage the benefits of both 3DGS and textured mesh rendering.

UV texture maps can be used to encode 3DGS parameters in a 2D ordered representation, that can be easily modeled with convolutional networks, while maintaining 3DMM-control \cite{xiang2024flashavatar, abdal2024gaussian, li2024animatablegaussian}.
This enables the end-to-end optimization of person-specific 3DGS head models \cite{teotia2024gaussianheads}.
Arguably, Relightable Gaussian Codec Avatars \cite{saito2024relightable}, demonstrated impressive results when trained on densely captured multi-view data, and enabled gaze and illumination control.
They have also been recently extended to a generalizable pipeline \cite{li2024uravatar} that can be even optimized on handheld device training data.
This generalizable framework can also be extended to training only on in-the-wild facial data, when paired with appropriate regularization  \cite{kirschstein2024gghead}.
To reduce the computational requirements, such convolutional 3DGS avatars can be distilled to smaller PCA models for real-time rendering \cite{zielonka2024gem, saito2024squeezeme}.
Finally, recent works show that generic 3DGS models pre-training on synthetic data enables fast and accurate adaptation to real data \cite{saunders2024gaspgaussianavatarssynthetic, zielonka2025synthetic}.
In common to this UV-space methods is their fixed number of Gaussian primitives, determined by the texels in the regressed Gaussian UV map.
Without an option for densifying (and pruning) Gaussians during training, resulting avatars often either lack geometric details for smaller UV maps, or are memory intensive for larger maps.
Our method instead significantly improves on visual quality with the proposed deformation model which retains the ability to adaptively densify more detailed regions.

\section{Method}

\subsection{Overview}

Given a set of high-resolution, multiview training images, our primary goal is to reconstruct a photorealistic representation of the captured human geometry and appearance, for high-fidelity rendering under controllable expressions and viewpoints. To do so, and inspired by related work in \refsec{sec:related_work}, we leverage a 3DMM to define and track an underlying mesh-based deformation model that serves as control signal for both training and animating our head model. 

The 3D Gaussians used for photorealistic rendering are defined relative to this underlying deformable mesh. Rather than anchoring Gaussians to a triangle~\cite{Qian2024gaussianavatars}, we instead define the 3D Gaussian positions in the continuous UVD space defined by the mesh's UV texture coordinates and displacement D along the local surface normal. This representation allows the Gaussians to freely move across different triangles during optimization, which contrasts with the greedy assignment method in~\cite{Qian2024gaussianavatars}. Our UVD-space optimization can thus recover from potentially suboptimal triangle assignments during densification, to increase modeling quality where most needed.

\begin{figure}[htb]
    \centering
    \includegraphics[width=\columnwidth]{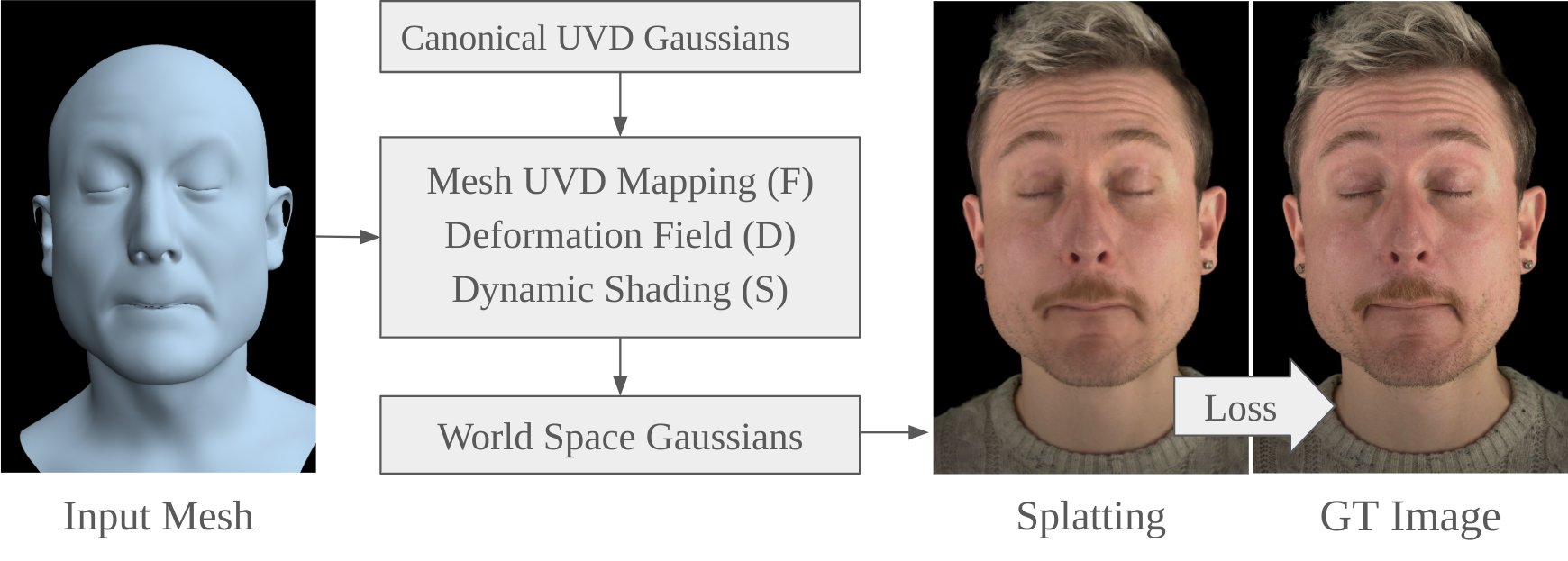}
    \caption{Our method learns a dense set of canonical 3D Gaussians within the continuous UVD tangent space of a given mesh model, essentially encoding a sparse volumetric texture. An expression-dependent (dynamic) appearance model is also learned to protorealistically render an arbitrary input mesh.}
    \label{fig:overview}
\end{figure}

However, the coarse nature of the above mesh-based head model most often leads to imprecise deformation estimates and, thus, loss of spatial detail. We therefore complement the model with a new, continuous UVD deformation field and expression-dependent dynamic shading to adjust the 3D Gaussian attributes for representing facial expressions, such as wrinkle deformation with associated darkening through ambient occlusion. As shown in our experimental results (\refsec{sec:results}), our overall model with these additional dynamic components can maintain sharp spatial detail (obtained via Gaussian densification) while also modeling skin deformation and transient facial features due to dynamic expressions (\eg, wrinkling). An overview of our method can be found in \reffig{fig:overview}.

\subsection{3D Gaussian Splatting}
\label{sec:preliminaries_3dgs}

3D Gaussian Splatting \cite{kerbl20233d} provides a fast reconstruction and rendering method, based on anisotropic 3D Gaussian primitives.
The 3D Gaussians $\mathbf{G} := \{ \boldsymbol{\mu},\mathbf{q}, \mathbf{s}, \mathbf{h}, \alpha \}$
are each parameterised by their location $\boldsymbol{\mu} \in \mathbb{R}^{3}$,
rotation quarternion $\mathbf{q} \in \mathbb{R}^{4}$,
scaling factor $\mathbf{s} \in \mathbb{R}^{3}$,
spherical harmonics parameters $\mathbf{h} \in \mathbb{R}^{16\times 3}$,
and opacity $\alpha \in \mathbb{R}$.
These correspond essentially to a parametric ellipse that is rasterized, 
with scaling matrix $\mathbf{S}$, 
rotation matrix $\mathbf{R}$ and therefore covariance matrix
$\mathbf{\Sigma} = \mathbf{RSS}^{\top}\mathbf{R}^{\top}$.
Given a view direction ${\bf v} \in \mathbb{R}^3$, The view-dependent RGB color of each Gaussian is computed as ${\bf c}({\bf v}) = {\bf b}_{SH}({\bf v}){\bf h}$, where ${\bf b}_{SH}({\bf v})$ is the spherical harmonics basis evaluated at ${\bf v}$. After splatting, the rendered pixel color is then a weighted blend of all the corresponding overlapping Gaussians, multiplied by their opacity.

\subsection{3DMM and Canonical UVD-space Gaussians}
\label{sec:3DMM}

Akin to other models such as FLAME~\cite{li2017flame}, our 3DMM defines a subspace of 3D head meshes with varying facial expressions and head poses for a given identity.
These meshes $\mathcal{M}_t$ are parameterized by expression parameters $\boldsymbol{\psi}_t \in \mathbb{R}^{313}$ and pose parameters $\boldsymbol{\theta}_t \in \mathbb{R}^{15}$, the latter capturing overall head translation as well as rotations of head, neck, and the two eyes. Note however that our model is not tied to our particular 3DMM implementation and can use, at training and test time, any sequence of 3D meshes $\mathcal{M}_t$, $t \in [1, T]$ that are in correspondence. We demonstrate this fact in \refsec{sec:results} by using head meshes provided with the Multiface dataset~\cite{wuu2022multiface}.

Rather than defining 3D Gaussian positions directly in 3D world space, our model defines {\em canonical} 3D Gaussians within the continuous UVD tangent space, shared by the subject's 3D meshes at training and test time. This representation allows us to establish a simple, continuous mapping between a Gaussian's location in UVD space ($\boldsymbol{\mu}_\text{uvd}$) and world space ($\boldsymbol{\mu}_\text{xyz}$),
\begin{align}
\label{eqn:mesh_mapping}
    \boldsymbol{\mu}_\text{xyz} = F(\boldsymbol{\mu}_\text{uvd}) \,.
\end{align}
Using traditional computer graphics, this function maps a UV coordinate to a 3D point on the corresponding mesh triangle, then uses the displacement coordinate D to offset the point along its normal (obtained from the triangle's vertices via barycentric interpolation). As a result, even if the canonical 3D Gaussian remains static in UVD space, it still moves, rotates and stretches in world space together with the underlying mesh surface. This full local transformation near a UVD point is captured by the Jacobian $J_F$ of $F(\cdot)$ at said point. We thus propose to use $J_F$ to propagate the deformation of the mesh surface to the Gaussian's covariance $\Sigma_\text{uvd}$ in UVD to obtain the covariance $\Sigma_\text{xyz}$ in world space:
\begin{equation}
\label{eqn:mesh_mapping_sigma}
    \Sigma_\text{xyz} = J_F \Sigma_\text{uvd} J_F^T .
\end{equation}
This model generalizes the Gaussian ``rigging'' by Qian \etal~\shortcite{Qian2024gaussianavatars}, which without explicit formalization does in fact use the Jacobian of their (scaled) rigid transformation. Our Jacobian-based approach is simpler to implement, as it can be computed using automatic differentiation using PyTorch. Furthermore, it also generalizes to any continuous 3D space mapping, without being tied to a particular 3DMM or deformation model. For example, it is able to handle non-rigid triangle deformations such as when facial skin stretches.
During optimization, canonical Gaussians are spawned, moved, and their covariance is $\Sigma$ reshaped within UVD space. Our Gaussian rigging method maintains the full solution space and allows Gaussians to move across triangles during optimization, if that leads to better densification and modeling. In contrast, GaussianAvatars \cite{Qian2024gaussianavatars} adopts a greedy and potentially suboptimal solution in which the binding of each Gaussian to a triangle cannot be undone. 

\subsection{UVD Deformation Field}

Continuous deformation fields have often been added to complement static (canonical) volumetric representations, turning the latter into deformable models \cite{Chen2024MonoGaussianAvatar,wang2024gaussianhead,yang2023deformable3dgs,yu2023cogs,Wu_2024_CVPR,lu2024gagaussian,park2021nerfies,Zielonka2022Insta}. Typically, these fields must learn to output a multidimensional parameter vector for a transformation that includes translation, rotation, and scaling components; such complex deformation fields often remain underconstrained (poorly supervised), and usually lead to poor generalization to unseen facial expressions.

We now revisit this approach and propose using a simple 3D translation field that is easier to train. Leveraging the Jacobian method in~\refeq{eqn:mesh_mapping_sigma}, we derive complete transformations from such simpler field, including rotation and stretching components; the shape of each 3D Gaussian thus simply deforms according with the continuous mapping from UVD to world space.

Here, our goal is to complement the coarse mesh-based deformation described above with a residual motion field that helps the overall model to better capture localized and subtle facial deformations, which in turn leads to modeling with a higher level of detail. Leveraging our Jacobian method above, we propose a deformation field $D(\cdot)$ that, given an input position $\boldsymbol{\mu}_\text{uvd}$ in UVD space, learns to output a residual 3D translation component in world space $\Delta \boldsymbol{\mu}_\text{xyz}$. The mapping in \refeq{eqn:mesh_mapping} then expands to
\begin{align}
\label{eqn:full_mapping}
\boldsymbol{\mu}_\text{xyz} = F(\boldsymbol{\mu}_\text{uvd}) + \Delta \boldsymbol{\mu}_\text{xyz} \,, \quad \text{where} \quad \Delta \boldsymbol{\mu}_\text{xyz} = D(\boldsymbol{\mu}_\text{uvd}; {\bf f}_\text{uv})\,.
\end{align}
This residual translation field is conditioned on a learned, local feature vector, ${\bf f}_\text{uv}$, that depends only on the Gaussian's UV coordinates and the underlying mesh deformation; ${\bf f}_\text{uv}$ is further detailed below, with the architecture of our deformation field. With this additional field, our \rev{Jacobian based covariance mapping in \refeq{eqn:mesh_mapping_sigma} becomes:}
\begin{equation}
\label{eqn:full_mapping_sigma}
    \Sigma_\text{xyz} = (J_F + J_D) \Sigma_\text{uvd} (J_F + J_D)^T = J_{(F+D)} \Sigma_\text{uvd} J_{(F+D)}^T \,,
\end{equation}
where $J_D$ is the new Jacobian of the deformation field relative to $\boldsymbol{\mu}_\text{uvd}$ only. \rev{The resulting Jacobian $J_{(F+D)} = J_F + J_D$ can then be directly computed via automatic differentiation.}

\begin{figure}[htb]
    \centering
    \includegraphics[width=0.8\columnwidth]{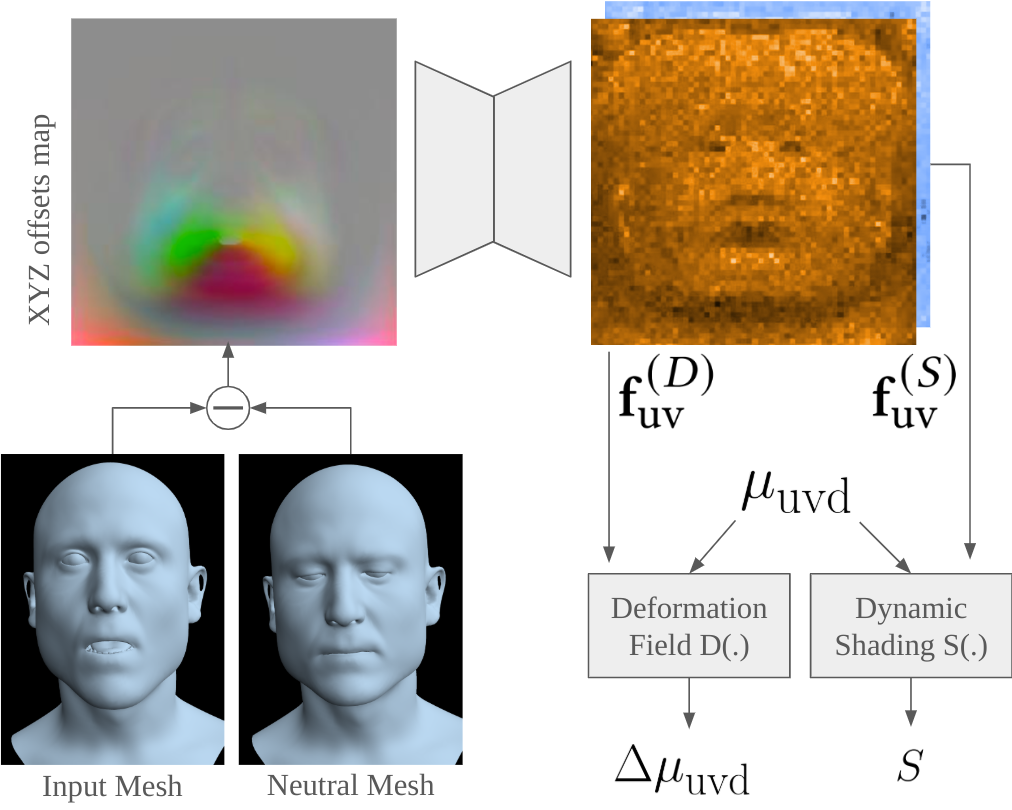}
    \caption{The UVD deformation field and dynamic shading components are modeled as MLPs that are conditioned on expression-dependent latent features learned on the UV texture grid by a shared U-Net encoder.}
    \label{fig:deformation_architecture}
\end{figure}

In contrast to the analytical mapping $F(\cdot)$, the deformation field $D(\cdot)$ is learned end-to-end and is comprised of two main components: ($i$) a convolutional U-Net, operating in the UV texture grid to output the local deformation codes ${\bf f}_\text{uv}$; and ($ii$) a shallow MLP network that implements $D(.)$ in \refeq{eqn:full_mapping}. This architecture is illustrated in~\reffig{fig:deformation_architecture}.
The input to the UV deformation U-Net is a texture of rasterized XYZ vertex coordinates in world space; these coordinates are represented as vertex displacements (offsets) relative to a similar texture obtained for a neutral face, using simple texelwise subtraction. To compensate for rigid head motion, we zero out the pose parameters for these displacements, leaving only the pure expressions. The U-Net is then trained to translate such offsets into a $N \times N \times 64$ latent texture from which the learned, local deformation codes ${\bf f}_\text{uv}$ can be queried with continuous UV coordinates via bilinear interpolation. This model thus allows us to decouple the texture dimension $N$ from the total number of Gaussians used for modeling (densification). In contrast, other methods \cite{li2024uravatar,teotia2024gaussianheads} that directly decode a texture of Gaussian attributes are limited to a fixed budget of at most $N^2$ Gaussians. For our experiments, we use a value of 256 for N.

The continuous deformation field $D(\cdot)$ is defined as a small MLP with two hidden layers, taking as input the position of a Gaussian in UVD space, $\boldsymbol{\mu}_\text{uvd}$ (with standard, sinusoidal positional encoding), and the corresponding input latent code ${\bf f}_\text{uv}$. As noted in Bai et al.~\shortcite{bai2023learning}, conditioning this MLP on such local expression code allows the model to adapt to localized deformations in the input 3D mesh. Additionally, in contrast to related work, our UVD deformation field leverages the Jacobian method above to simplify its output, making the learning task simpler.
Finally, as human eyes are rigid, we exclude any Gaussians placed on the eyeballs from being deformed by the deformation field.

\subsection{Expression-Dependent Dynamic Colors}

While the deformation model above can adjust the position and covariance of the 3D Gaussians for a given expression, the view-dependent RGB color of each Gaussian, ${\bf c}({\bf v}) \in \mathbb{R}^3$, remains as an expression-agnostic ``canonical'' attribute. To model transient appearance features such as darkening within skin wrinkles, we now derive an additional network component, $S(\cdot)$, to provide each Gaussian with an \rev{expression-dependent dynamic shading}.

To render our expression-dependent Gaussians in world space, we define a dynamic \rev{shading} model ${\bf c}_e({\bf v}, {\bf f}_\text{uv})$ that is also a function of the latent facial expression code ${\bf f}_\text{uv}$,
\begin{align}
     {\bf c}_e({\bf v}, {\bf f}_\text{uv}) = {\bf c}({\bf v}) S(\boldsymbol{\mu}_\text{uvd}; {\bf f}_\text{uv})\,.
\end{align}
The canonical color term ${\bf c}({\bf v})$ has the standard Spherical Harmonics encoding ${\bf h}$, while the dynamic \rev{monochromatic} shading term $S(\cdot) \in \mathbb{R}$ is modeled as a small MLP with two hidden layers.

Empirically, we found it beneficial to duplicate the last layer of the deformation U-Net so that a dedicated (split) set of latents ${\bf f}_\text{uv}^{(S)}$ is learned for dynamic colors, as shown in~\reffig{fig:deformation_architecture}. This change allows us to better accommodate different regularization (smoothness) constraints on these two dynamic components. As detailed next, we found it advantageous to have a spatially smooth deformation field, without smoothing the dynamic shading network.

\subsection{Training Losses and Regularization}

Given input video frames captured in multiple views, we assume that a deformable 3D face mesh model (with fixed topology and UV map) has been tracked over these frames. We then proceed to train our new dynamic, photorealistic appearance model by minimizing an overall rendering loss, between captured $\mathcal{I}$ and rendered $\hat{\mathcal{I}}$ images, which is also subject to regularization constraints as described next.

Following related work, we utilize a combination of $\mathcal{L}_1$, $\mathcal{L}_{D-SSIM}$, and $\mathcal{L}_{VGG}$ losses as our primary objective, 
\begin{align}
     \mathcal{L} = \lambda_1\mathcal{L}_1 + \lambda_{D-SSIM}\mathcal{L}_{D-SSIM} + \lambda_{VGG}\mathcal{L}_{VGG} +  \lambda_{J_{D}}\mathcal{L}_{J_{D}} \,,
\end{align}
with $\lambda_1=0.8$ and $\lambda_{D-SSIM}=0.2$, as originally proposed in~\cite{kerbl20233d}. The perceptual VGG loss is defined as:
\begin{align}
    \mathcal{L}_{VGG} = ||F_{VGG}(\mathcal{I}) - F_{VGG}(\hat{\mathcal{I}})||
\end{align}
where $F_{VGG}(\cdot)$ denotes the output feature of the first four layers of a pre-trained VGG network \cite{8578166}. While this perceptual loss helps reconstruct finer detail exceptionally well, our experiments show that it may also interfere with the optimization of the UVD deformation field during initial iterations. We thus follow a coarse-to-fine training strategy, in which we initially set $\lambda_{VGG} = 0$. After 50\% of the training is done, when the model can start to focus on learning finer details, we set $\lambda_{VGG}=0.01$ and exponentially increase towards a final value of 0.1. This leads to better estimation of non-rigid motion across video frames, and sharper appearance detail as a result.

We also find it beneficial to constrain the UVD deformation field to be initially smoother. We enforce smoothness by penalizing the spatial derivatives of this field,
\begin{align}
    \mathcal{L}_{J_{D}} = \sum_{s_{\text{uvd}}} \big\lVert J_{D}(s_{\text{uvd}}) \big\rVert_{\text{F}}^{2}
\end{align}
where $\lVert\cdot\rVert_{\text{F}}$ denotes the Frobenius norm and $J_{D}(s_{\text{uvd}})$ is the field's Jacobian evaluated at 100K randomly sampled UVD points $s_{\text{uvd}}$, with $UV \in [0,1]$ and $D \in [-50, 200]$mm. The goal here is to have the entire deformation field be spatially smooth, not just at the Gaussian means. We initially set $\lambda_\text{smooth}=1.0$, which then exponentially decays to $0.1$ halfway during training.
Also, following \citet{li2024uravatar}, we restrict the size of each Gaussian to within a prescribed range, having standard deviations within $[5mm, 0.02mm]$.

We train our method over 60K total iterations on an NVIDIA H100, taking 10 to 16 hours. Our PyTorch implementation uses an Adam optimizer\cite{adam} with learning rates set as in \citet{yang2023deformable3dgs}, decaying over the 60K iterations.

\subsection{Initialization}

We initialize the weights of the deformation and shading MLPs with zeros, which leads to a null initial field and uniform dynamic shading at 1.0 (using a sigmoid activation multiplied with 2).
To initialize the canonical UVD Gaussians, we uniformly sample 500K random means $\boldsymbol{\mu}_\text{uvd}$ and, for the remaining attributes, we follow the procedure in Kerbl et al.~\shortcite{kerbl20233d}. Because some Gaussian may be initialized outside valid areas of the UV plane, we project these means onto the nearest mesh triangle.

As \citet{yang2023deformable3dgs}, we "warm up" the initial canonical 3D Gaussians by first optimizing them without the residual UVD deformation field, and only begin optimizing the field after obtaining an initial canonical volume (after 10K iterations). Unlike \citet{yang2023deformable3dgs}, our method can warm up for substantially longer by leveraging the coarse deformation given by the underlying 3DMM mesh.

\subsection{Adaptive Densification}
Unlike many other deformable face avatar approaches \cite{xiang2024flashavatar,Chen2024MonoGaussianAvatar,li2024uravatar,xu2023gaussianheadavatar}, we retain the adaptive densification from 3DGS \cite{kerbl20233d} and their thresholding -- 
we clone small Gaussians, and split large Gaussians with large view-space gradients.
As these thresholds are computed using screen-space gradients, the densification algorithm works without any changes.
More specifically, although densification thresholds are computed in world-space, 
splitting and cloning occurs in UVD space, which is itself a continuous 3D space.
Finally, we set an upper bound of 4 million Gaussians to avoid running out of memory.
In this manner, we can benefit by the 3DMM guidance and UV ordered topology,
while being able to reconstruct and track fine facial and hair details,
which vary significantly among individuals.
\section{Multiview Image Datasets}

We trained and evaluated our new method against baselines on two multiview image datasets: (1) the publicly available Multiface dataset \cite{wuu2022multiface}, and (2) our own dataset captured in-studio with the facial performance setup described next.

Our multiview, calibrated camera rig comprises 13 26MP cameras with 50mm lenses, evenly distributed over the frontal hemisphere and tightly framed to capture the subject's head down to their shoulders. Constant, uniform illumination is provided by 14 small area lights evenly placed in front of the subject. For our experiments, we captured 8 subjects, each performing a set of prescribed facial motions that included facial expressions of emotion, localized muscle activation (action units) and different lip shapes (visemes), for a total of 44 short video clips (at 30fps) per subject. On each video frame, dense multiview stereo~\cite{gu2020cascade} was run to reconstruct a 3D mesh, followed by registration of a template 3D mesh topology, which was done by tracking our own 3D morphable model (similar to FLAME~\cite{li2017flame}) across the input videos. For training a head avatar model for each subject, we sampled 400 frames uniformly from the input video takes and downsampled the image dimensions by half, to $3072 \times 2048$ pixels (due to GPU memory limitations).
Finally, we run a matting algorithm \cite{pandey2021total} to mask out background pixels.

For Multiface, we selected a subset of 40 cameras (excluding cameras behind the head and directly above). We trained and tested on subjects with IDs 5372021, 6795937, 8870559, 002643814. For each subject, we also uniformly sampled 400 frames for training. Instead of tracking a 3DMM, we use the 3D meshes provided with the dataset for all evaluated methods, including ours. We train at the original resolution of $2048 \times 1334$ pixels without downsampling. 

For both datasets, we consider test sequences where the subject goes through multiple expressions over the course of a few seconds.

\begin{table*}[p]
\caption{Quantitative comparison on our dataset; metrics evaluated on the full head. \hlredcaption{Best}, \hlorangecaption{second best}, \hlyellowcaption{third best} scores are highlighted.}
\resizebox{\textwidth}{!}{%
\centering
\begin{tabular}{cccccccc} 
      \toprule
      & & \multicolumn{3}{c}{Test Reenactment} & \multicolumn{3}{c}{Novel View} \\
      \hline
        & Landmark $\downarrow$ & LPIPS $\downarrow$ & SSIM $\uparrow$ & PSNR $\uparrow$ & LPIPS $\downarrow$ & SSIM $\uparrow$ & PSNR $\uparrow$ \\
      \hline
      No Warp & 34.0 $\pm$ 32.7 & 0.169 $\pm$ 0.045 & \hlyellow{0.779 $\pm$ 0.054} & 24.3 $\pm$ 1.3 & 0.139 $\pm$ 0.042 & 0.828 $\pm$ 0.040 & 26.1 $\pm$ 1.7 \\ 
      No Shading & \hlyellow{33.5 $\pm$ 31.0} & \hlyellow{0.155 $\pm$ 0.053} & 0.772 $\pm$ 0.060 & 24.0 $\pm$ 1.2 & \hlyellow{0.128 $\pm$ 0.048} & 0.847 $\pm$ 0.037 & 26.3 $\pm$ 1.4 \\ 
      No Network & 38.2 $\pm$ 33.8 & 0.180 $\pm$ 0.050 & 0.769 $\pm$ 0.058 & 23.6 $\pm$ 1.4 & 0.150 $\pm$ 0.044 & 0.818 $\pm$ 0.043 & 25.1 $\pm$ 1.7\\ 
      No Densify & 122.7 $\pm$ 147 & 0.274 $\pm$ 0.072 & 0.738 $\pm$ 0.053 & 21.3 $\pm$ 1.5 & 0.211 $\pm$ 0.061 & 0.802 $\pm$ 0.044 & 24.9 $\pm$ 1.5 \\
      MLP & 34.4 $\pm$ 31.7 & 0.167 $\pm$ 0.049 & 0.778 $\pm$ 0.058 & \hlorange{24.5 $\pm$ 1.3} & 0.130 $\pm$ 0.049 & 0.833 $\pm$ 0.043 & 26.1 $\pm$ 1.5 \\
      No VGG & 37.1 $\pm$ 34.8 & 0.178 $\pm$ 0.052 & \hlred{0.787 $\pm$ 0.032} & \hlred{24.5 $\pm$ 1.2} & 0.148 $\pm$ 0.052 & \hlorange{0.851 $\pm$ 0.037} & \hlorange{26.8 $\pm$ 1.5} \\
      No Triangle Updates &  \hlorange{31.7 $\pm$ 29.1} & \hlorange{0.153 $\pm$ 0.016} & 0.777 $\pm$ 0.061 & 24.3 $\pm$ 1.4 & \hlorange{0.123 $\pm$ 0.048} & \hlred{0.853 $\pm$ 0.039} & \hlred{27.0 $\pm$ 1.6} \\
      \hline
      GaussianAvatars & 71.7 $\pm$ 67.7 & 0.249 $\pm$ 0.060 & 0.773 $\pm$ 0.050 & 24.1 $\pm$ 1.5 & 0.201 $\pm$ 0.060 & 0.813 $\pm$ 0.043 & 25.4 $\pm$ 1.6 \\
      RGCA  & 116.9 $\pm$ 95.6  & 0.212 $\pm$ 0.055 & 0.730 $\pm$ 0.058 & 23.0 $\pm$ 1.2 & 0.177 $\pm$ 0.055 & 0.823 $\pm$ 0.060 & 24.4 $\pm$ 5.1 \\
      GHA & 1421 $\pm$ 1625  & 0.229 $\pm$ 0.075 & 0.669 $\pm$ 0.064 & 15.5 $\pm$ 2.3 & 0.180 $\pm$ 0.080 & 0.725 $\pm$ 0.073 & 17.1 $\pm$ 2.1 \\
      MVP & 208.0 $\pm$ 122.  & 0.268 $\pm$ 0.071 & 0.748 $\pm$ 0.051 & 23.1 $\pm$ 1.3 & 0.226 $\pm$ 0.080 & 0.800 $\pm$ 0.049 & 26.0 $\pm$ 1.3 \\
      \hline
      Ours  & \hlred{31.7 $\pm$ 28.6} & \hlred{0.150 $\pm$ 0.055} & \hlorange{0.779 $\pm$ 0.061} & \hlyellow{24.4 $\pm$ 1.2} & \hlred{0.123 $\pm$ 0.045}  & \hlyellow{0.848 $\pm$ 0.038} &  \hlyellow{26.7 $\pm$ 1.5} \\ 
      Ours (200K GS.) & 49.0 $\pm$ 26.1 & 0.207 $\pm$ 0.022 & 0.770 $\pm$ 0.029 & 24.3 $\pm$ 1.4 & 0.163 $\pm$ 0.013 & 0.807 $\pm$ 0.024 & 25.7 $\pm$ 1.3 \\
      \bottomrule
\end{tabular}%
}
\label{tab:test1}
\end{table*}
\section{Results}
\label{sec:results}

\begin{figure*}[p]
    \centering
    \makebox[0.166\textwidth][c]{GT}\makebox[0.166\textwidth][c]{Ours}\makebox[0.166\textwidth][c]{RGCA}\makebox[0.166\textwidth][c]{GA}\makebox[0.166\textwidth][c]{MVP}\makebox[0.166\textwidth][c]{GHA}
    
    \vspace{1mm}

    \includegraphics[width=0.996\textwidth]{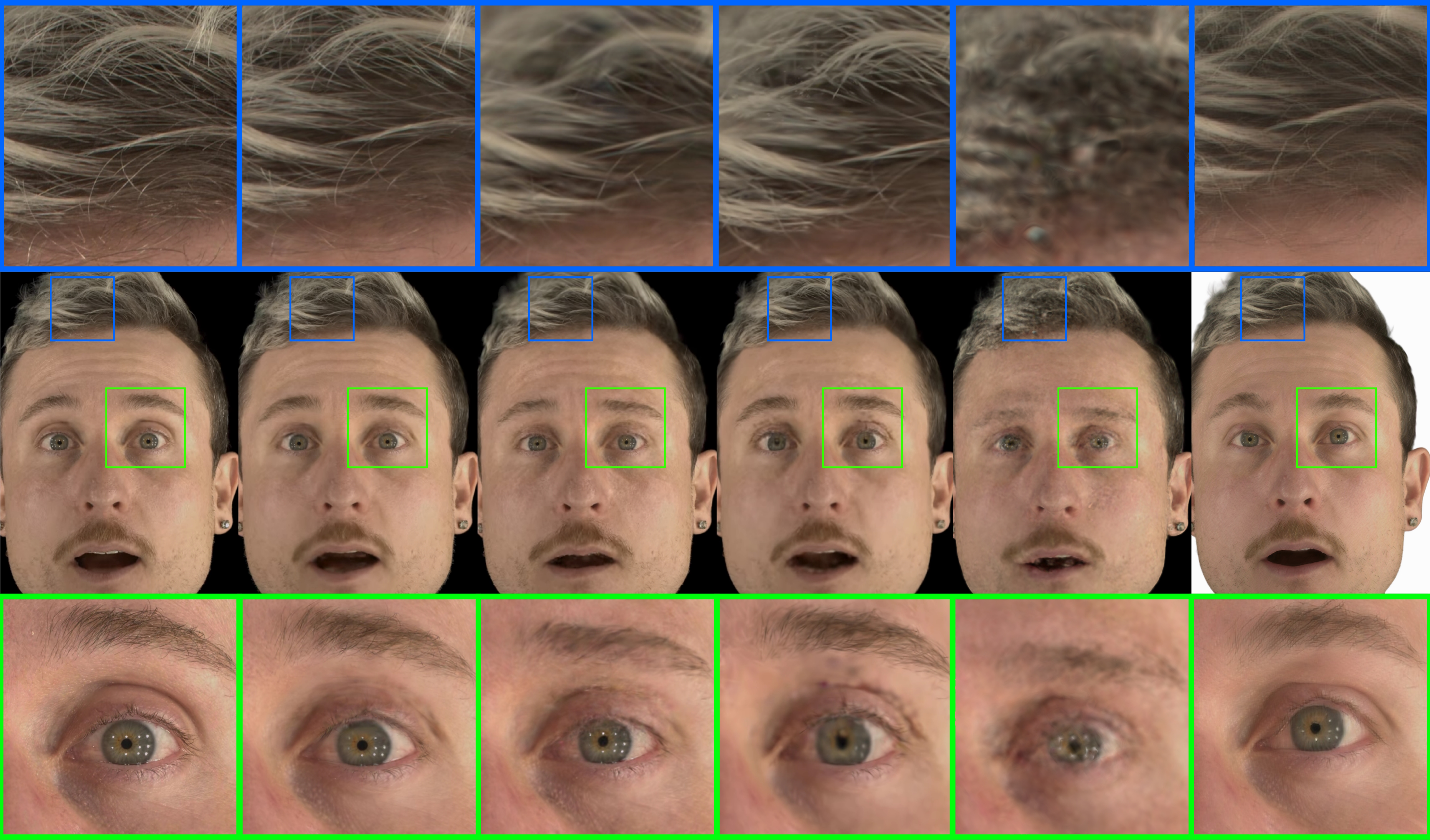}    
    
    \caption{Our method manages to extrapolate better when compared to RGCA, GA, MVP and GHA. Thanks to the warpfield and densification we can more faithfuly capture hair and brows.}

\end{figure*}

In this section, we demonstrate facial performance reenactment as well as novel view synthesis. For each subject, we evaluate on test frames from three training views, using three cameras distributed horizontally in front of the face. We also evaluate novel view synthesis quality on images rendered on 3 holdout camera views, using training expression. We strongly encourage readers to view the supplementary materials for full resolution videos.

We compare LPIPS, SSIM and PSNR, computed relative to known ground-truth images. To evaluate expression alignment, we use an off-the-shelf landmark detector, and estimate the mean-squared difference between keypoints predicted on the grouth truth versus rendered image. To improve consistency, we only use landmarks on the eye, mouth and nose.

\subsection{Comparisons with the state of the art.}

We compare our method against \rev{four representative baselines, GaussianAvatars (GA)~\cite{Qian2024gaussianavatars}, GaussianHeadAvatars (GHA)~\cite{xu2023gaussianheadavatar}, RGCA~\cite{saito2024relightable},  and MVP~\cite{Lombardi21}} that focus on high-quality modeling from in-studio data. Thus, we do not include baselines that target monocular reconstruction in the wild, with lower modeling quality.

As RGCA is designed to be relightable, we adjust it to be compatible with a static lighting dataset without access to environment lighting information. To do this, we replace their specular and light direction dependent SH formulation with the original view dependent SH formulation. Other than that, we run their code as is, using the same geometry as used with our method. As our multiview dataset does not have accurate projected textures, we instead use the same vertex delta map used as an input for our U-Net. \rev{MVP, using a similar architecture, is trained using the same inputs, but as it models only a static environment, requires no further changes.}

GaussianAvatars is run on the same geometry as our method and RGCA. As Multiface provides tracked meshes, rather than a 3DMM, we disable pose optimization and input the mesh directly.

\rev{Finally, GHA uses supervision based on proximity to 3D keypoints. Therefore, we used their provided data processing pipeline.}

\begin{figure*}
    \centering
    \makebox[0.166\textwidth][c]{GT}\makebox[0.166\textwidth][c]{Ours}\makebox[0.166\textwidth][c]{RGCA}\makebox[0.166\textwidth][c]{GA}\makebox[0.166\textwidth][c]{MVP}\makebox[0.166\textwidth][c]{GHA}

    \vspace{1mm}

    \includegraphics[width=0.996\textwidth]{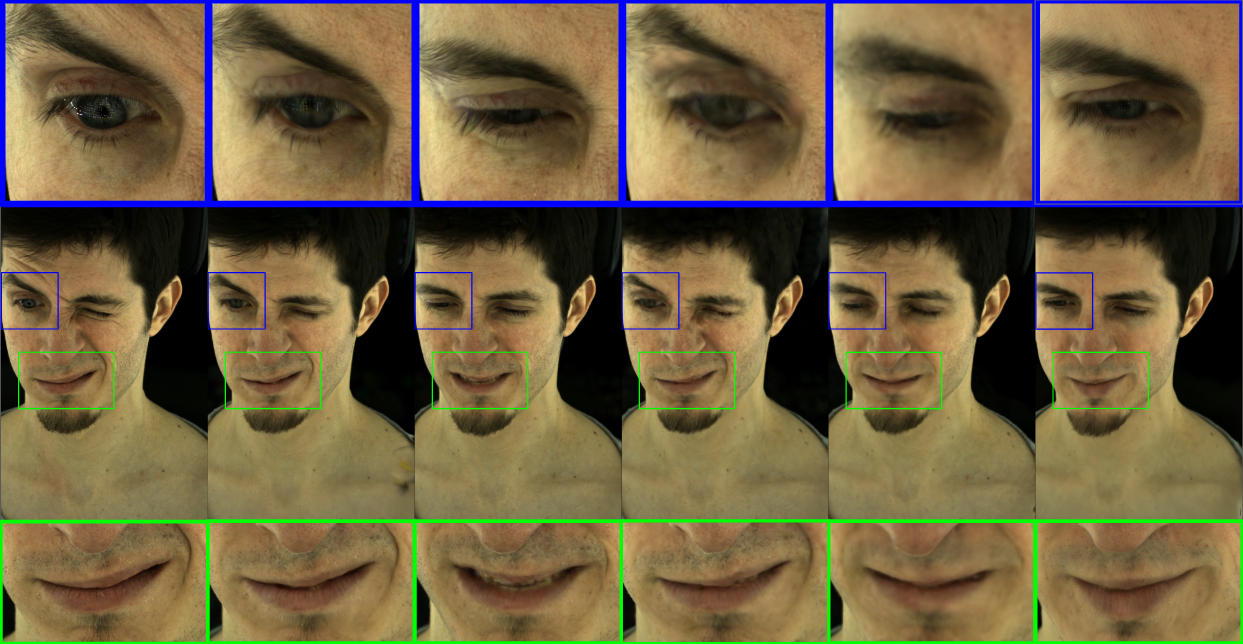}
    
    \vspace{-0.3mm}

    \includegraphics[width=0.996\textwidth]{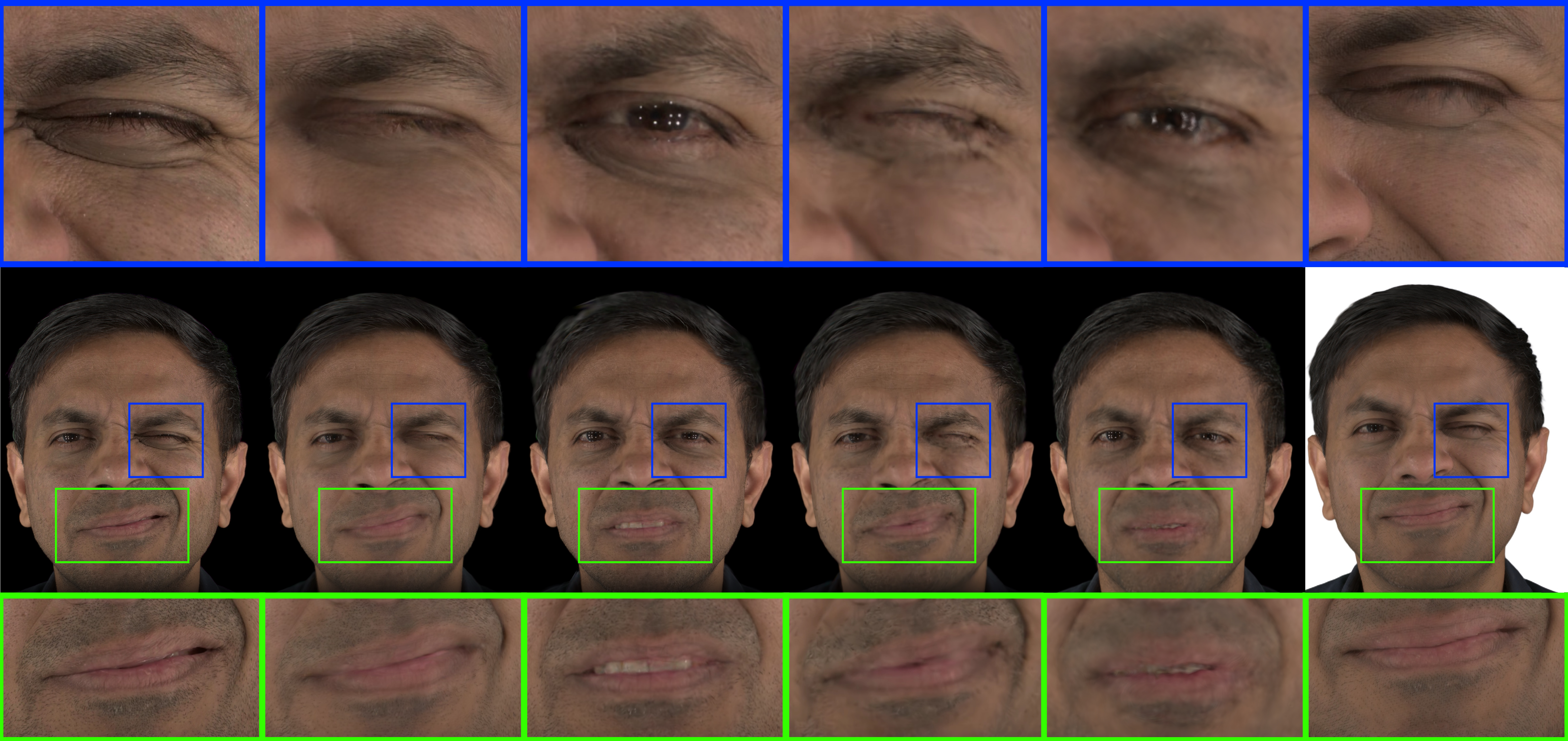}
    \vspace{-1mm}
    \caption{Our method manages to extrapolate better when compared to RGCA, GA, MVP and GHA. Thanks to the warpfield and densification we can more faithfuly capture hair and brows (top-left). RGCA struggles to recreate expressions unseen during training (top, subject from Multiface\cite{wuu2022multiface}) due to its architecture. GA extrapolates the coarse shape better, but suffers from reduced spatial detail and exhibits artefacts since it doesn't explicitly model dynamic appearance (i.e. lower-right). We encourage readers to zoom in further to see more details}
    \label{fig:results_1}
\end{figure*}

As \reffig{fig:results_1} shows, RGCA significantly underperforms in expression preservation (alignment). This is likely due to auto-encoding vertex locations: on novel test data, the auto-encoded expressions are reproduced less accurately, especially with more limited training data. Their convolutional decoder is also limited to a fixed budget of Gaussians (about 1 million), being unable to run adaptive densification. As such, their results are blurrier and suffer from ``transparency'' artifacts in novel views due to the representation not being dense enough (visible when the camera moves, as shown in our supplementary material). \rev{MVP, being limited to fixed voxel primtives, has similar issues and overall performs worse than RGCA.}

GaussianAvatars performs well at lower resolutions, and is mostly able to reconstruct expressions. However, when zoomed in, their method fails to reconstruct fine details and has numerous artifacts. \rev{Despite using adaptive densification, their method typically still uses less than 200K Gaussians. For a fairer comparison, we rerun our method with a similar upper limit, and show that we outperform them even with a similar gaussian budget.} Please refer to the several side-by-side videos in our supplementary material.

\rev{GHA suffers from noticeable misalignments as well as aliasing artifacts. GHA was notably the only method which could not use alternate geometry, and their custom reconstruction uses a purely keypoint based 3DMM fitting approach, resulting in overall worse fits due to a lack of photometric consistency. Furthermore, we note that both our dataset and the multiface dataset have much crisper images compared to the NeRFSemble dataset used by GHA. As a result, the previously already present aliasing artifacts are significantly more accentuated and noticeable.}

\subsection{Ablations}

Table \ref{tab:test1} shows ablations on our various design choices and settings to demonstrate their contribution to the method.

\subsubsection{Warp Field and Shading model}

\begin{figure}[t]
    \centering
    \includegraphics[width=\patchwidth]{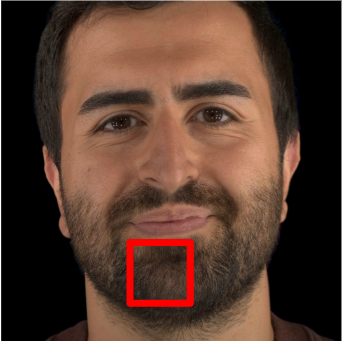}\hspace{0.03\columnwidth}\includegraphics[width=\patchwidth]{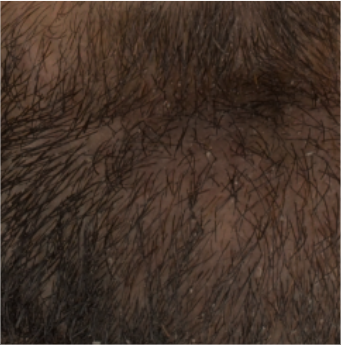}\hspace{0.03\columnwidth}\includegraphics[width=\patchwidth]{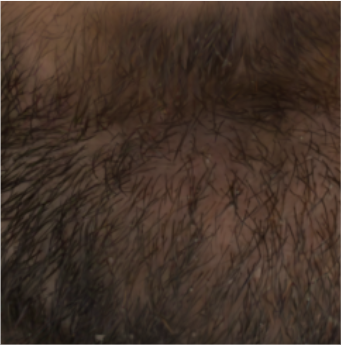}

    \vspace{-1.0mm}

    \makebox[\patchwidth][c]{\small GT}\hspace{0.03\columnwidth}\makebox[\patchwidth][c]{\small GT}\hspace{0.03\columnwidth}\makebox[\patchwidth][c]{\small Ours}
    
    \vspace{
    0.5mm}
    
    \includegraphics[width=\patchwidth]{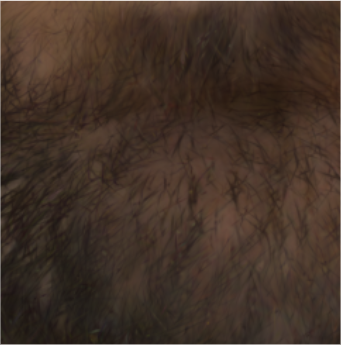}\hspace{0.03\columnwidth}\includegraphics[width=\patchwidth]{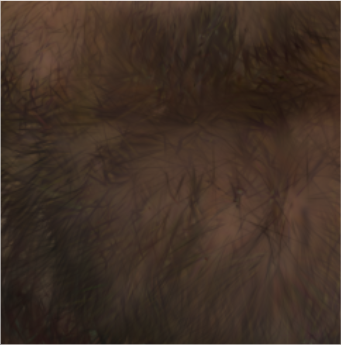}
    
    \vspace{-1.0mm}

    \makebox[\patchwidth][c]{\small MLP}\hspace{0.03\columnwidth}\makebox[\patchwidth][c]{\small No Warp}

    \vspace{-2mm}

    \caption{We compare zoomed in patches of the red highlighted area.}
    \label{fig:warps}
\end{figure}

\begin{figure}
	\centering
	\includegraphics[trim={0 1.3cm 0 0},clip, width=\columnwidth]{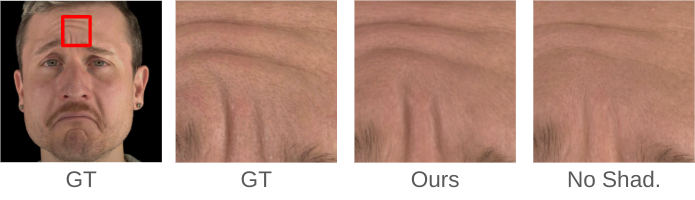}
	
	\makebox[0.25\columnwidth][c]{GT}\makebox[0.25\columnwidth][c]{GT}\makebox[0.25\columnwidth][c]{Ours}\makebox[0.25\columnwidth][c]{No Shading}
	
	\vspace{-1mm}
	\caption{Close-up view of the highlighted patch, showing that our dynamic shading leads to better modeling of skin wrinkles.}
	\label{fig:noshading}
\end{figure}

The deformation and shading nets are inherently ambiguous and capable of compensating each other. We thus also evaluated our method trained without shading (No Shading), without deformation (No Warp), and without both (No Network). As \reffig{fig:noshading} shows, the shading network is critical to reconstructing wrinkles well. Similarly, without a deformation network, our method is unable to reconstruct finer details nearly as well, \reffig{fig:warps}. We also ablate our choice of a hybrid CNN/MLP approach, instead of a deep MLP: we replaced our UNET with an 8 layer CNN encoder which produces a single latent code for the entire expression, and use it to drive a positionally encoded MLP, similar to \citet{yang2023deformable3dgs} (MLP). However, due to the comparative expensiveness of a deep MLP, we can only train 1.5 million Gaussians with the same amount of GPU memory. \reffig{fig:warps} shows this is insufficient to best capture finer details.

\subsubsection{Triangle Updates}

We show the benefit of allowing Gaussians to move across triangles by comparing with the binding inheritance strategy from \citet{Qian2024gaussianavatars} (No triangle updates). Without the ability to switch Gaussians between triangles, artifacts appear in many blinking poses, as seen in~\reffig{fig:misc}.

\subsubsection{Densification}

Critical to our method is the ability to adaptively densify. As seen in Figure \ref{fig:misc}, without densification (No Densify), our method dramatically drops in quality.

\subsubsection{VGG loss}

\begin{figure}[h]
	\centering
	\includegraphics[width=\patchwidth]{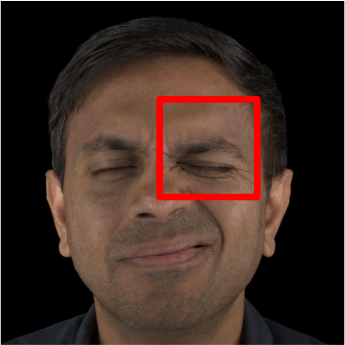}\hspace{0.03\columnwidth}\includegraphics[width=\patchwidth]{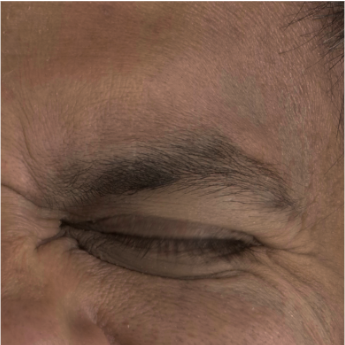}\hspace{0.03\columnwidth}\includegraphics[width=\patchwidth]{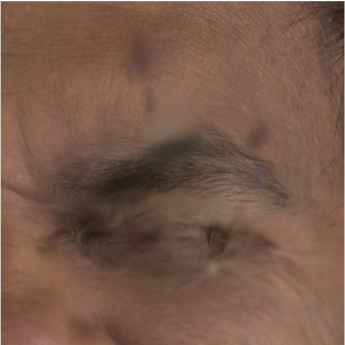}
	
	\vspace{-1.0mm}
	
	\makebox[\patchwidth][c]{\small GT}\hspace{0.03\columnwidth}\makebox[\patchwidth][c]{\small GT}\hspace{0.03\columnwidth}\makebox[\patchwidth][c]{\small No Triangle Updates}
	
	\vspace{
		0.5mm}
	
	\includegraphics[width=\patchwidth]{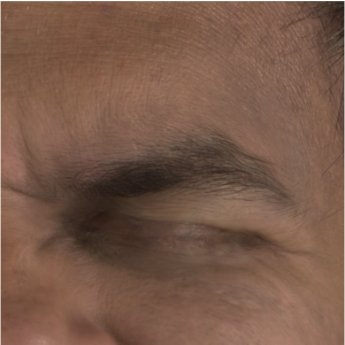}\hspace{0.03\columnwidth}\includegraphics[width=\patchwidth]{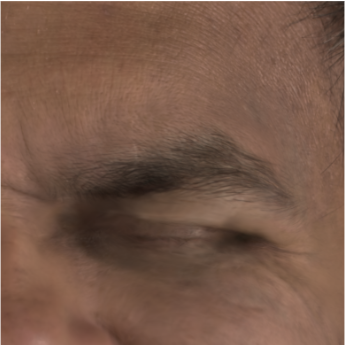}\hspace{0.03\columnwidth}\includegraphics[width=\patchwidth]{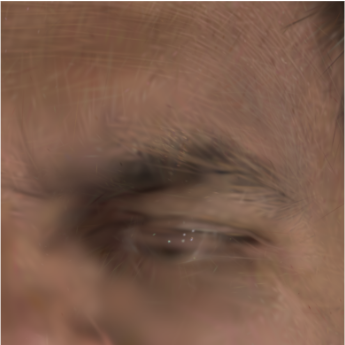}
	
	\vspace{-1.0mm}
	
	\makebox[\patchwidth][c]{\small Ours}\hspace{0.03\columnwidth}\makebox[\patchwidth][c]{\small No VGG}\hspace{0.03\columnwidth}\makebox[\patchwidth][c]{\small No Densify}
	
	\vspace{-2mm}
	\caption{We compare zoomed in patches of the red highlighted area.}
	\label{fig:misc}
\end{figure}

We note that neither GaussianAvatars nor RGCA use a perceptual VGG loss. Therefore, for fairness, we run our method without it (No VGG), and show that though we lose quality (Figure~\ref{fig:misc})  we still outperform state-of-the-art methods without it.

\section{Conclusion}

We present a novel method for reconstructing photoreal head avatars that can be rendered at high resolution with controllable expressions and viewpoints. Our hybrid approach includes ($i$) a simple ``rigging'' model within the continuous tangent space of a facial 3DMM, where canonical Gaussians move freely across triangles for non-greedy optimization; and ($ii$) a network for expression-dependent deformation and dynamic shading that is agnostic to the number of 3D Gaussians, allowing for densification where most needed and for capturing both static and transient skin features in high detail. In comparison to baselines representing the state-of-the-art, our results show better reconstruction of both overall expression and level of detail.

Our method is however not without limitations: it relies on tracked 3D meshes for training, which are less accurate on areas such as the teeth, tongue, and mouth interior in general; loss of detail and reconstruction artifacts can thus be observed on these areas (which are also more poorly constrained by the training data). In addition, our dynamic shading term cannot change the chromaticity of each Gaussian and thus cannot account for dynamic skin appearance effects due to blood flow. Future work could investigate improving tongue and teeth reconstruction, as well as more comprehensive dynamic appearance effects.

\begin{figure*}
    \centering
    \makebox[0.1525\textwidth][c]{GT}\makebox[0.1525\textwidth][c]{Ours}\makebox[0.1525\textwidth][c]{RGCA}\makebox[0.1525\textwidth][c]{GA}\makebox[0.1525\textwidth][c]{MVP}\makebox[0.1525\textwidth][c]{GHA}

    \vspace{1mm}

    \includegraphics[width=0.915\textwidth]{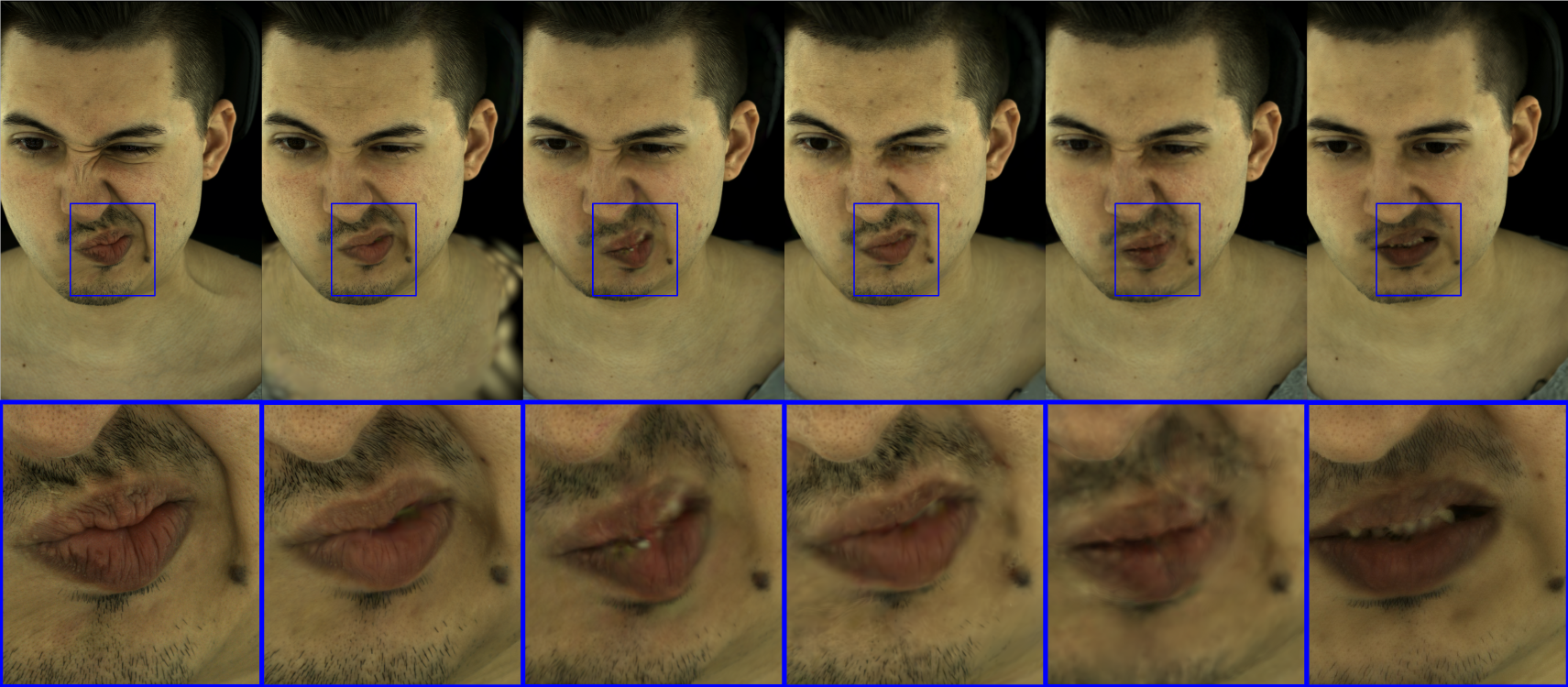}
    
    \vspace{-0.3mm}
    
    \includegraphics[width=0.915\textwidth]{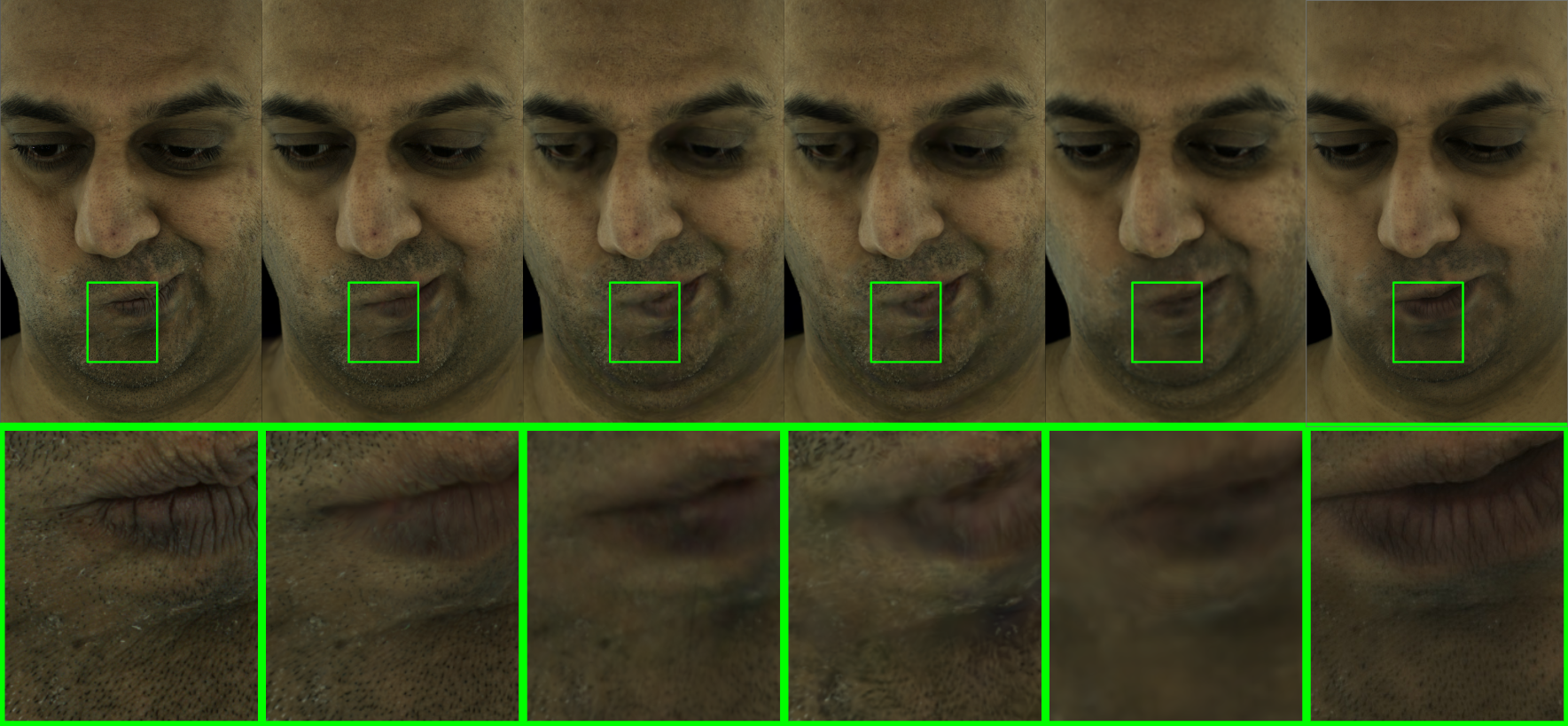}
    
    \vspace{-0.3mm}
    
    \includegraphics[width=0.915\textwidth]{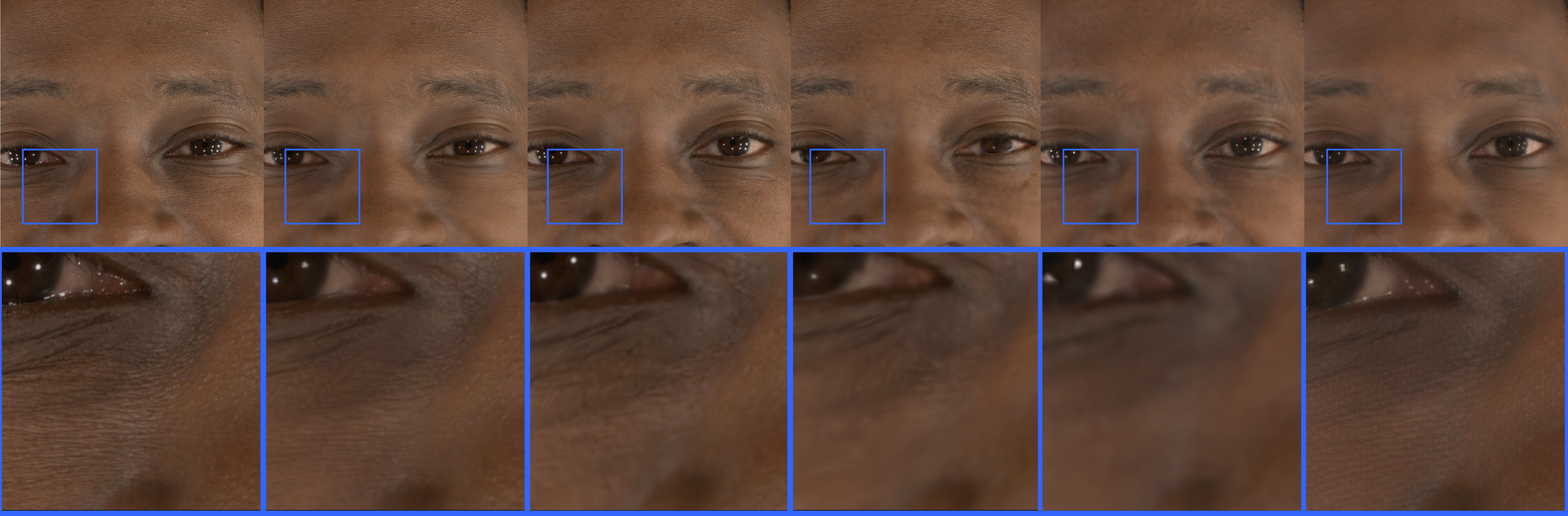}
    
    \vspace{-1mm}
    
    \caption{In comparison to RGCA, GA, MVP and GHA our approach has significantly higher fidelity on details, such as stubble (top, middle, subjects from Multiface \cite{wuu2022multiface}) or small wrinkles (bottom).}
    \label{fig:results_2}
\end{figure*}

\newpage

\bibliographystyle{ACM-Reference-Format}
\bibliography{bibliography.bib}


\begin{thebibliography}{61}


\ifx \showCODEN    \undefined \def \showCODEN     #1{\unskip}     \fi
\ifx \showISBNx    \undefined \def \showISBNx     #1{\unskip}     \fi
\ifx \showISBNxiii \undefined \def \showISBNxiii  #1{\unskip}     \fi
\ifx \showISSN     \undefined \def \showISSN      #1{\unskip}     \fi
\ifx \showLCCN     \undefined \def \showLCCN      #1{\unskip}     \fi
\ifx \shownote     \undefined \def \shownote      #1{#1}          \fi
\ifx \showarticletitle \undefined \def \showarticletitle #1{#1}   \fi
\ifx \showURL      \undefined \def \showURL       {\relax}        \fi
\providecommand\bibfield[2]{#2}
\providecommand\bibinfo[2]{#2}
\providecommand\natexlab[1]{#1}
\providecommand\showeprint[2][]{arXiv:#2}

\bibitem[Abdal et~al\mbox{.}(2024)]%
        {abdal2024gaussian}
\bibfield{author}{\bibinfo{person}{Rameen Abdal}, \bibinfo{person}{Wang Yifan},
  \bibinfo{person}{Zifan Shi}, \bibinfo{person}{Yinghao Xu},
  \bibinfo{person}{Ryan Po}, \bibinfo{person}{Zhengfei Kuang},
  \bibinfo{person}{Qifeng Chen}, \bibinfo{person}{Dit-Yan Yeung}, {and}
  \bibinfo{person}{Gordon Wetzstein}.} \bibinfo{year}{2024}\natexlab{}.
\newblock \showarticletitle{Gaussian shell maps for efficient {3D} human
  generation}. In \bibinfo{booktitle}{\emph{Conference on Computer Vision and
  Pattern Recognition (CVPR)}}. \bibinfo{publisher}{CVF / IEEE},
  \bibinfo{pages}{9441--9451}.
\newblock


\bibitem[Bai et~al\mbox{.}(2023)]%
        {bai2023learning}
\bibfield{author}{\bibinfo{person}{Ziqian Bai}, \bibinfo{person}{Feitong Tan},
  \bibinfo{person}{Zeng Huang}, \bibinfo{person}{Kripasindhu Sarkar},
  \bibinfo{person}{Danhang Tang}, \bibinfo{person}{Di Qiu},
  \bibinfo{person}{Abhimitra Meka}, \bibinfo{person}{Ruofei Du},
  \bibinfo{person}{Mingsong Dou}, \bibinfo{person}{Sergio Orts-Escolano},
  {et~al\mbox{.}}} \bibinfo{year}{2023}\natexlab{}.
\newblock \showarticletitle{Learning personalized high quality volumetric head
  avatars from monocular {RGB} videos}. In \bibinfo{booktitle}{\emph{Conference
  on Computer Vision and Pattern Recognition (CVPR)}}. \bibinfo{publisher}{CVF
  / IEEE}, \bibinfo{pages}{16890--16900}.
\newblock


\bibitem[Blanz and Vetter(1999)]%
        {blanz1999}
\bibfield{author}{\bibinfo{person}{Volker Blanz} {and} \bibinfo{person}{Thomas
  Vetter}.} \bibinfo{year}{1999}\natexlab{}.
\newblock \showarticletitle{A Morphable Model for the Synthesis of {3D} Faces}.
  In \bibinfo{booktitle}{\emph{SIGGRAPH}}. \bibinfo{pages}{187--194}.
\newblock


\bibitem[Booth et~al\mbox{.}(2016)]%
        {booth20163d}
\bibfield{author}{\bibinfo{person}{James Booth}, \bibinfo{person}{Anastasios
  Roussos}, \bibinfo{person}{Stefanos Zafeiriou}, \bibinfo{person}{Allan
  Ponniah}, {and} \bibinfo{person}{David Dunaway}.}
  \bibinfo{year}{2016}\natexlab{}.
\newblock \showarticletitle{A {3D} morphable model learnt from 10,000 faces}.
  In \bibinfo{booktitle}{\emph{Conference on Computer Vision and Pattern
  Recognition (CVPR)}}. \bibinfo{publisher}{CVF / IEEE},
  \bibinfo{pages}{5543--5552}.
\newblock


\bibitem[Chan et~al\mbox{.}(2022)]%
        {chan2022efficient}
\bibfield{author}{\bibinfo{person}{Eric~R Chan}, \bibinfo{person}{Connor~Z
  Lin}, \bibinfo{person}{Matthew~A Chan}, \bibinfo{person}{Koki Nagano},
  \bibinfo{person}{Boxiao Pan}, \bibinfo{person}{Shalini De~Mello},
  \bibinfo{person}{Orazio Gallo}, \bibinfo{person}{Leonidas~J Guibas},
  \bibinfo{person}{Jonathan Tremblay}, \bibinfo{person}{Sameh Khamis},
  {et~al\mbox{.}}} \bibinfo{year}{2022}\natexlab{}.
\newblock \showarticletitle{Efficient geometry-aware {3D} generative
  adversarial networks}. In \bibinfo{booktitle}{\emph{Conference on Computer
  Vision and Pattern Recognition (CVPR)}}. \bibinfo{publisher}{CVF / IEEE},
  \bibinfo{pages}{16123--16133}.
\newblock


\bibitem[Chen et~al\mbox{.}(2024)]%
        {Chen2024MonoGaussianAvatar}
\bibfield{author}{\bibinfo{person}{Yufan Chen}, \bibinfo{person}{Lizhen Wang},
  \bibinfo{person}{Qijing Li}, \bibinfo{person}{Hongjiang Xiao},
  \bibinfo{person}{Shengping Zhang}, \bibinfo{person}{Hongxun Yao}, {and}
  \bibinfo{person}{Yebin Liu}.} \bibinfo{year}{2024}\natexlab{}.
\newblock \showarticletitle{{MonoGaussianAvatar}: {M}onocular Gaussian
  Point-based Head Avatar}. In \bibinfo{booktitle}{\emph{SIGGRAPH Conference
  Papers (SA)}}. \bibinfo{publisher}{{ACM}}, \bibinfo{pages}{58}.
\newblock


\bibitem[Debevec(2012)]%
        {debevec2012light}
\bibfield{author}{\bibinfo{person}{Paul Debevec}.}
  \bibinfo{year}{2012}\natexlab{}.
\newblock \showarticletitle{The light stages and their applications to
  photoreal digital actors}.
\newblock \bibinfo{journal}{\emph{Transactions on Graphics, (Proc. SIGGRAPH
  Asia)}} \bibinfo{volume}{2}, \bibinfo{number}{4} (\bibinfo{year}{2012}),
  \bibinfo{pages}{1--6}.
\newblock


\bibitem[Dhamo et~al\mbox{.}(2024)]%
        {Dhamo2024headgas}
\bibfield{author}{\bibinfo{person}{Helisa Dhamo}, \bibinfo{person}{Yinyu Nie},
  \bibinfo{person}{Arthur Moreau}, \bibinfo{person}{Jifei Song},
  \bibinfo{person}{Richard Shaw}, \bibinfo{person}{Yiren Zhou}, {and}
  \bibinfo{person}{Eduardo P{\'{e}}rez{-}Pellitero}.}
  \bibinfo{year}{2024}\natexlab{}.
\newblock \showarticletitle{{HeadGaS}: {R}eal-Time Animatable Head Avatars via
  {3D} Gaussian Splatting}. In \bibinfo{booktitle}{\emph{European Conference on
  Computer Vision (ECCV)}}, Vol.~\bibinfo{volume}{15060}.
  \bibinfo{publisher}{Springer}, \bibinfo{pages}{459--476}.
\newblock


\bibitem[Egger et~al\mbox{.}(2020)]%
        {egger20203d}
\bibfield{author}{\bibinfo{person}{Bernhard Egger}, \bibinfo{person}{William~AP
  Smith}, \bibinfo{person}{Ayush Tewari}, \bibinfo{person}{Stefanie Wuhrer},
  \bibinfo{person}{Michael Zollhoefer}, \bibinfo{person}{Thabo Beeler},
  \bibinfo{person}{Florian Bernard}, \bibinfo{person}{Timo Bolkart},
  \bibinfo{person}{Adam Kortylewski}, \bibinfo{person}{Sami Romdhani},
  {et~al\mbox{.}}} \bibinfo{year}{2020}\natexlab{}.
\newblock \showarticletitle{{3D} morphable face models—past, present, and
  future}.
\newblock \bibinfo{journal}{\emph{Transactions on Graphics (TOG)}}
  \bibinfo{volume}{39}, \bibinfo{number}{5} (\bibinfo{year}{2020}),
  \bibinfo{pages}{1--38}.
\newblock


\bibitem[Gafni et~al\mbox{.}(2021)]%
        {Gafni2021nerface}
\bibfield{author}{\bibinfo{person}{Guy Gafni}, \bibinfo{person}{Justus Thies},
  \bibinfo{person}{Michael Zollh{\"o}fer}, {and} \bibinfo{person}{Matthias
  Nie{\ss}ner}.} \bibinfo{year}{2021}\natexlab{}.
\newblock \showarticletitle{Dynamic Neural Radiance Fields for Monocular {4D}
  Facial Avatar Reconstruction}. In \bibinfo{booktitle}{\emph{Conference on
  Computer Vision and Pattern Recognition (CVPR)}}. \bibinfo{publisher}{CVF /
  IEEE}, \bibinfo{pages}{8649--8658}.
\newblock


\bibitem[Gecer et~al\mbox{.}(2019)]%
        {gecer2019ganfit}
\bibfield{author}{\bibinfo{person}{Baris Gecer}, \bibinfo{person}{Stylianos
  Ploumpis}, \bibinfo{person}{Irene Kotsia}, {and} \bibinfo{person}{Stefanos
  Zafeiriou}.} \bibinfo{year}{2019}\natexlab{}.
\newblock \showarticletitle{{GANFIT}: {G}enerative adversarial network fitting
  for high fidelity {3D} face reconstruction}. In
  \bibinfo{booktitle}{\emph{Conference on Computer Vision and Pattern
  Recognition (CVPR)}}. \bibinfo{publisher}{CVF / IEEE},
  \bibinfo{pages}{1155--1164}.
\newblock


\bibitem[Giebenhain et~al\mbox{.}(2024)]%
        {giebenhain2024npga}
\bibfield{author}{\bibinfo{person}{Simon Giebenhain}, \bibinfo{person}{Tobias
  Kirschstein}, \bibinfo{person}{Martin R{\"{u}}nz}, \bibinfo{person}{Lourdes
  Agapito}, {and} \bibinfo{person}{Matthias Nie{\ss}ner}.}
  \bibinfo{year}{2024}\natexlab{}.
\newblock \showarticletitle{{NPGA}: {N}eural Parametric Gaussian Avatars}. In
  \bibinfo{booktitle}{\emph{SIGGRAPH Conference Papers (SA)}}.
  \bibinfo{pages}{127:1--127:11}.
\newblock


\bibitem[Goodfellow et~al\mbox{.}(2014)]%
        {goodfellow2014generative}
\bibfield{author}{\bibinfo{person}{Ian Goodfellow}, \bibinfo{person}{Jean
  Pouget-Abadie}, \bibinfo{person}{Mehdi Mirza}, \bibinfo{person}{Bing Xu},
  \bibinfo{person}{David Warde-Farley}, \bibinfo{person}{Sherjil Ozair},
  \bibinfo{person}{Aaron Courville}, {and} \bibinfo{person}{Yoshua Bengio}.}
  \bibinfo{year}{2014}\natexlab{}.
\newblock \showarticletitle{Generative adversarial nets}.
\newblock \bibinfo{journal}{\emph{Advances in Neural Information Processing
  Systems (NeurIPS)}}  \bibinfo{volume}{27} (\bibinfo{year}{2014}).
\newblock


\bibitem[Gu et~al\mbox{.}(2020)]%
        {gu2020cascade}
\bibfield{author}{\bibinfo{person}{Xiaodong Gu}, \bibinfo{person}{Zhiwen Fan},
  \bibinfo{person}{Siyu Zhu}, \bibinfo{person}{Zuozhuo Dai},
  \bibinfo{person}{Feitong Tan}, {and} \bibinfo{person}{Ping Tan}.}
  \bibinfo{year}{2020}\natexlab{}.
\newblock \showarticletitle{Cascade Cost Volume for High-Resolution Multi-View
  Stereo and Stereo Matching}. In \bibinfo{booktitle}{\emph{Conference on
  Computer Vision and Pattern Recognition (CVPR)}}.
  \bibinfo{pages}{2495--2504}.
\newblock


\bibitem[Hong et~al\mbox{.}(2022)]%
        {hong2021headnerf}
\bibfield{author}{\bibinfo{person}{Yang Hong}, \bibinfo{person}{Bo Peng},
  \bibinfo{person}{Haiyao Xiao}, \bibinfo{person}{Ligang Liu}, {and}
  \bibinfo{person}{Juyong Zhang}.} \bibinfo{year}{2022}\natexlab{}.
\newblock \showarticletitle{{HeadNeRF}: {A} Real-time NeRF-based Parametric
  Head Model}. In \bibinfo{booktitle}{\emph{Conference on Computer Vision and
  Pattern Recognition (CVPR)}}. \bibinfo{publisher}{CVF / IEEE},
  \bibinfo{pages}{20342--20352}.
\newblock


\bibitem[Karras et~al\mbox{.}(2020)]%
        {karras2020analyzing}
\bibfield{author}{\bibinfo{person}{Tero Karras}, \bibinfo{person}{Samuli
  Laine}, \bibinfo{person}{Miika Aittala}, \bibinfo{person}{Janne Hellsten},
  \bibinfo{person}{Jaakko Lehtinen}, {and} \bibinfo{person}{Timo Aila}.}
  \bibinfo{year}{2020}\natexlab{}.
\newblock \showarticletitle{Analyzing and improving the image quality of
  stylegan}. In \bibinfo{booktitle}{\emph{Conference on Computer Vision and
  Pattern Recognition (CVPR)}}. \bibinfo{publisher}{CVF / IEEE},
  \bibinfo{pages}{8110--8119}.
\newblock


\bibitem[Kerbl et~al\mbox{.}(2023)]%
        {kerbl20233d}
\bibfield{author}{\bibinfo{person}{Bernhard Kerbl}, \bibinfo{person}{Georgios
  Kopanas}, \bibinfo{person}{Thomas Leimk{\"u}hler}, {and}
  \bibinfo{person}{George Drettakis}.} \bibinfo{year}{2023}\natexlab{}.
\newblock \showarticletitle{3d gaussian splatting for real-time radiance field
  rendering.}
\newblock \bibinfo{journal}{\emph{Transactions on Graphics (TOG)}}
  \bibinfo{volume}{42}, \bibinfo{number}{4} (\bibinfo{year}{2023}),
  \bibinfo{pages}{139--1}.
\newblock


\bibitem[Kingma and Ba(2014)]%
        {adam}
\bibfield{author}{\bibinfo{person}{Diederik~P Kingma} {and}
  \bibinfo{person}{Jimmy Ba}.} \bibinfo{year}{2014}\natexlab{}.
\newblock \showarticletitle{{Adam: A Method for Stochastic Optimization}}.
\newblock \bibinfo{journal}{\emph{ArXiv}} (\bibinfo{year}{2014}).
\newblock
\href{https://doi.org/10.48550/arXiv.1412.6980}{doi:\nolinkurl{10.48550/arXiv.1412.6980}}


\bibitem[Kirschstein et~al\mbox{.}(2024)]%
        {kirschstein2024gghead}
\bibfield{author}{\bibinfo{person}{Tobias Kirschstein}, \bibinfo{person}{Simon
  Giebenhain}, \bibinfo{person}{Jiapeng Tang}, \bibinfo{person}{Markos
  Georgopoulos}, {and} \bibinfo{person}{Matthias Nie{\ss}ner}.}
  \bibinfo{year}{2024}\natexlab{}.
\newblock \showarticletitle{{GGHead}: {F}ast and Generalizable {3D} Gaussian
  Heads}. In \bibinfo{booktitle}{\emph{SIGGRAPH Asia Conference Papers (SA)}}.
  \bibinfo{publisher}{{ACM}}, \bibinfo{pages}{126:1--126:11}.
\newblock


\bibitem[Lattas et~al\mbox{.}(2020)]%
        {lattas2020avatarme}
\bibfield{author}{\bibinfo{person}{Alexandros Lattas},
  \bibinfo{person}{Stylianos Moschoglou}, \bibinfo{person}{Baris Gecer},
  \bibinfo{person}{Stylianos Ploumpis}, \bibinfo{person}{Vasileios
  Triantafyllou}, \bibinfo{person}{Abhijeet Ghosh}, {and}
  \bibinfo{person}{Stefanos Zafeiriou}.} \bibinfo{year}{2020}\natexlab{}.
\newblock \showarticletitle{{AvatarMe}: {R}ealistically Renderable 3D Facial
  Reconstruction" in-the-wild"}. In \bibinfo{booktitle}{\emph{Conference on
  Computer Vision and Pattern Recognition (CVPR)}}. \bibinfo{publisher}{CVF /
  IEEE}, \bibinfo{pages}{760--769}.
\newblock


\bibitem[Lattas et~al\mbox{.}(2023)]%
        {lattas2023fitme}
\bibfield{author}{\bibinfo{person}{Alexandros Lattas},
  \bibinfo{person}{Stylianos Moschoglou}, \bibinfo{person}{Stylianos Ploumpis},
  \bibinfo{person}{Baris Gecer}, \bibinfo{person}{Jiankang Deng}, {and}
  \bibinfo{person}{Stefanos Zafeiriou}.} \bibinfo{year}{2023}\natexlab{}.
\newblock \showarticletitle{{FitMe}: {D}eep photorealistic {3D} morphable model
  avatars}. In \bibinfo{booktitle}{\emph{Conference on Computer Vision and
  Pattern Recognition (CVPR)}}. \bibinfo{publisher}{CVF / IEEE},
  \bibinfo{pages}{8629--8640}.
\newblock


\bibitem[Li et~al\mbox{.}(2024b)]%
        {shellnerf2024}
\bibfield{author}{\bibinfo{person}{Gengyan Li}, \bibinfo{person}{Kripasindhu
  Sarkar}, \bibinfo{person}{Abhimitra Meka}, \bibinfo{person}{Marcel Buehler},
  \bibinfo{person}{Franziska Mueller}, \bibinfo{person}{Paulo Gotardo},
  \bibinfo{person}{Otmar Hilliges}, {and} \bibinfo{person}{Thabo Beeler}.}
  \bibinfo{year}{2024}\natexlab{b}.
\newblock \showarticletitle{{ShellNeRF: Learning a Controllable High-resolution
  Model of the Eye and Periocular Region}}.
\newblock \bibinfo{journal}{\emph{Computer Graphics Forum}}
  (\bibinfo{year}{2024}).
\newblock
\showISSN{1467-8659}
\href{https://doi.org/10.1111/cgf.15041}{doi:\nolinkurl{10.1111/cgf.15041}}


\bibitem[Li et~al\mbox{.}(2024a)]%
        {li2024uravatar}
\bibfield{author}{\bibinfo{person}{Junxuan Li}, \bibinfo{person}{Chen Cao},
  \bibinfo{person}{Gabriel Schwartz}, \bibinfo{person}{Rawal Khirodkar},
  \bibinfo{person}{Christian Richardt}, \bibinfo{person}{Tomas Simon},
  \bibinfo{person}{Yaser Sheikh}, {and} \bibinfo{person}{Shunsuke Saito}.}
  \bibinfo{year}{2024}\natexlab{a}.
\newblock \showarticletitle{{URAvatar}: {U}niversal Relightable Gaussian Codec
  Avatars}. In \bibinfo{booktitle}{\emph{SIGGRAPH Conference Papers (SA)}}.
  \bibinfo{pages}{128:1--128:11}.
\newblock


\bibitem[Li et~al\mbox{.}(2020)]%
        {li2020learning}
\bibfield{author}{\bibinfo{person}{Ruilong Li}, \bibinfo{person}{Karl Bladin},
  \bibinfo{person}{Yajie Zhao}, \bibinfo{person}{Chinmay Chinara},
  \bibinfo{person}{Owen Ingraham}, \bibinfo{person}{Pengda Xiang},
  \bibinfo{person}{Xinglei Ren}, \bibinfo{person}{Pratusha Prasad},
  \bibinfo{person}{Bipin Kishore}, \bibinfo{person}{Jun Xing}, {et~al\mbox{.}}}
  \bibinfo{year}{2020}\natexlab{}.
\newblock \showarticletitle{Learning formation of physically-based face
  attributes}. In \bibinfo{booktitle}{\emph{Conference on Computer Vision and
  Pattern Recognition (CVPR)}}. \bibinfo{publisher}{CVF / IEEE},
  \bibinfo{pages}{3410--3419}.
\newblock


\bibitem[Li et~al\mbox{.}(2017)]%
        {li2017flame}
\bibfield{author}{\bibinfo{person}{Tianye Li}, \bibinfo{person}{Timo Bolkart},
  \bibinfo{person}{Michael~J Black}, \bibinfo{person}{Hao Li}, {and}
  \bibinfo{person}{Javier Romero}.} \bibinfo{year}{2017}\natexlab{}.
\newblock \showarticletitle{Learning a model of facial shape and expression
  from 4D scans.}
\newblock  \bibinfo{volume}{36}, \bibinfo{number}{6} (\bibinfo{year}{2017}),
  \bibinfo{pages}{194--1}.
\newblock


\bibitem[Li et~al\mbox{.}(2024c)]%
        {li2024animatablegaussian}
\bibfield{author}{\bibinfo{person}{Zhe Li}, \bibinfo{person}{Zerong Zheng},
  \bibinfo{person}{Lizhen Wang}, {and} \bibinfo{person}{Yebin Liu}.}
  \bibinfo{year}{2024}\natexlab{c}.
\newblock \showarticletitle{{Animatable Gaussians}: {L}earning Pose-Dependent
  Gaussian Maps for High-Fidelity Human Avatar Modeling}. In
  \bibinfo{booktitle}{\emph{Conference on Computer Vision and Pattern
  Recognition (CVPR)}}. \bibinfo{publisher}{CVF / IEEE},
  \bibinfo{pages}{19711--19722}.
\newblock


\bibitem[Lombardi et~al\mbox{.}(2021)]%
        {Lombardi21}
\bibfield{author}{\bibinfo{person}{Stephen Lombardi}, \bibinfo{person}{Tomas
  Simon}, \bibinfo{person}{Gabriel Schwartz}, \bibinfo{person}{Michael
  Zollhoefer}, \bibinfo{person}{Yaser Sheikh}, {and} \bibinfo{person}{Jason
  Saragih}.} \bibinfo{year}{2021}\natexlab{}.
\newblock \showarticletitle{Mixture of Volumetric Primitives for Efficient
  Neural Rendering}.
\newblock \bibinfo{journal}{\emph{ACM Trans. Graph.}} \bibinfo{volume}{40},
  \bibinfo{number}{4}, Article \bibinfo{articleno}{59} (\bibinfo{date}{jul}
  \bibinfo{year}{2021}), \bibinfo{numpages}{13}~pages.
\newblock
\showISSN{0730-0301}
\href{https://doi.org/10.1145/3450626.3459863}{doi:\nolinkurl{10.1145/3450626.3459863}}


\bibitem[Lu et~al\mbox{.}(2024)]%
        {lu2024gagaussian}
\bibfield{author}{\bibinfo{person}{Zhicheng Lu}, \bibinfo{person}{Xiang Guo},
  \bibinfo{person}{Le Hui}, \bibinfo{person}{Tianrui Chen},
  \bibinfo{person}{Ming Yang}, \bibinfo{person}{Xiao Tang},
  \bibinfo{person}{Feng Zhu}, {and} \bibinfo{person}{Yuchao Dai}.}
  \bibinfo{year}{2024}\natexlab{}.
\newblock \showarticletitle{{3D} Geometry-aware Deformable Gaussian Splatting
  for Dynamic View Synthesis}. In \bibinfo{booktitle}{\emph{Conference on
  Computer Vision and Pattern Recognition (CVPR)}}. \bibinfo{publisher}{CVF /
  IEEE}, \bibinfo{pages}{8900--8910}.
\newblock


\bibitem[Luiten et~al\mbox{.}(2024)]%
        {luiten2024dynamic}
\bibfield{author}{\bibinfo{person}{Jonathon Luiten}, \bibinfo{person}{Georgios
  Kopanas}, \bibinfo{person}{Bastian Leibe}, {and} \bibinfo{person}{Deva
  Ramanan}.} \bibinfo{year}{2024}\natexlab{}.
\newblock \showarticletitle{Dynamic {3D} {G}aussians: {T}racking by persistent
  dynamic view synthesis}. In \bibinfo{booktitle}{\emph{International
  Conference on 3D Vision (3DV)}}. IEEE, \bibinfo{pages}{800--809}.
\newblock


\bibitem[Luo et~al\mbox{.}(2021)]%
        {luo2021normalized}
\bibfield{author}{\bibinfo{person}{Huiwen Luo}, \bibinfo{person}{Koki Nagano},
  \bibinfo{person}{Han-Wei Kung}, \bibinfo{person}{Qingguo Xu},
  \bibinfo{person}{Zejian Wang}, \bibinfo{person}{Lingyu Wei},
  \bibinfo{person}{Liwen Hu}, {and} \bibinfo{person}{Hao Li}.}
  \bibinfo{year}{2021}\natexlab{}.
\newblock \showarticletitle{Normalized avatar synthesis using {StyleGAN} and
  perceptual refinement}. In \bibinfo{booktitle}{\emph{Conference on Computer
  Vision and Pattern Recognition (CVPR)}}. \bibinfo{publisher}{CVF / IEEE},
  \bibinfo{pages}{11662--11672}.
\newblock


\bibitem[Mildenhall et~al\mbox{.}(2020)]%
        {mildenhall2021nerf}
\bibfield{author}{\bibinfo{person}{Ben Mildenhall}, \bibinfo{person}{Pratul~P.
  Srinivasan}, \bibinfo{person}{Matthew Tancik}, \bibinfo{person}{Jonathan~T.
  Barron}, \bibinfo{person}{Ravi Ramamoorthi}, {and} \bibinfo{person}{Ren Ng}.}
  \bibinfo{year}{2020}\natexlab{}.
\newblock \showarticletitle{{NeRF}: {R}epresenting Scenes as Neural Radiance
  Fields for View Synthesis}. In \bibinfo{booktitle}{\emph{European Conference
  on Computer Vision (ECCV)}}, \bibfield{editor}{\bibinfo{person}{Andrea
  Vedaldi}, \bibinfo{person}{Horst Bischof}, \bibinfo{person}{Thomas Brox},
  {and} \bibinfo{person}{Jan{-}Michael Frahm}} (Eds.),
  Vol.~\bibinfo{volume}{12346}. \bibinfo{publisher}{Springer},
  \bibinfo{pages}{405--421}.
\newblock


\bibitem[Nagano et~al\mbox{.}(2018)]%
        {nagano2018pagan}
\bibfield{author}{\bibinfo{person}{Koki Nagano}, \bibinfo{person}{Jaewoo Seo},
  \bibinfo{person}{Jun Xing}, \bibinfo{person}{Lingyu Wei},
  \bibinfo{person}{Zimo Li}, \bibinfo{person}{Shunsuke Saito},
  \bibinfo{person}{Aviral Agarwal}, \bibinfo{person}{Jens Fursund},
  \bibinfo{person}{Hao Li}, \bibinfo{person}{Richard Roberts}, {et~al\mbox{.}}}
  \bibinfo{year}{2018}\natexlab{}.
\newblock \showarticletitle{{paGAN}: {R}eal-time avatars using dynamic
  textures.}
\newblock \bibinfo{journal}{\emph{Transactions on Graphics (TOG)}}
  \bibinfo{volume}{37}, \bibinfo{number}{6} (\bibinfo{year}{2018}),
  \bibinfo{pages}{258}.
\newblock


\bibitem[Pandey et~al\mbox{.}(2021)]%
        {pandey2021total}
\bibfield{author}{\bibinfo{person}{Rohit Pandey}, \bibinfo{person}{Sergio~Orts
  Escolano}, \bibinfo{person}{Chloe Legendre}, \bibinfo{person}{Christian
  Haene}, \bibinfo{person}{Sofien Bouaziz}, \bibinfo{person}{Christoph
  Rhemann}, \bibinfo{person}{Paul Debevec}, {and} \bibinfo{person}{Sean
  Fanello}.} \bibinfo{year}{2021}\natexlab{}.
\newblock \showarticletitle{Total relighting: learning to relight portraits for
  background replacement}.
\newblock \bibinfo{journal}{\emph{ACM Transactions on Graphics (TOG)}}
  \bibinfo{volume}{40}, \bibinfo{number}{4} (\bibinfo{year}{2021}),
  \bibinfo{pages}{1--21}.
\newblock


\bibitem[Papantoniou et~al\mbox{.}(2023)]%
        {papantoniou2023relightify}
\bibfield{author}{\bibinfo{person}{Foivos~Paraperas Papantoniou},
  \bibinfo{person}{Alexandros Lattas}, \bibinfo{person}{Stylianos Moschoglou},
  {and} \bibinfo{person}{Stefanos Zafeiriou}.} \bibinfo{year}{2023}\natexlab{}.
\newblock \showarticletitle{Relightify: Relightable 3d faces from a single
  image via diffusion models}. In \bibinfo{booktitle}{\emph{Conference on
  Computer Vision and Pattern Recognition (CVPR)}}. \bibinfo{publisher}{CVF /
  IEEE}, \bibinfo{pages}{8806--8817}.
\newblock


\bibitem[Park et~al\mbox{.}(2021)]%
        {park2021nerfies}
\bibfield{author}{\bibinfo{person}{Keunhong Park}, \bibinfo{person}{Utkarsh
  Sinha}, \bibinfo{person}{Jonathan~T. Barron}, \bibinfo{person}{Sofien
  Bouaziz}, \bibinfo{person}{Dan~B Goldman}, \bibinfo{person}{Steven~M. Seitz},
  {and} \bibinfo{person}{Ricardo Martin-Brualla}.}
  \bibinfo{year}{2021}\natexlab{}.
\newblock \showarticletitle{Nerfies: {D}eformable Neural Radiance Fields}.
\newblock \bibinfo{journal}{\emph{International Conference on Computer Vision
  (ICCV)}} (\bibinfo{year}{2021}).
\newblock


\bibitem[Paysan et~al\mbox{.}(2009)]%
        {paysan20093d}
\bibfield{author}{\bibinfo{person}{Pascal Paysan}, \bibinfo{person}{Reinhard
  Knothe}, \bibinfo{person}{Brian Amberg}, \bibinfo{person}{Sami Romdhani},
  {and} \bibinfo{person}{Thomas Vetter}.} \bibinfo{year}{2009}\natexlab{}.
\newblock \showarticletitle{A {3D} face model for pose and illumination
  invariant face recognition}. In \bibinfo{booktitle}{\emph{International
  conference on advanced video and signal based surveillance}}. IEEE,
  \bibinfo{pages}{296--301}.
\newblock


\bibitem[Qian et~al\mbox{.}(2024)]%
        {Qian2024gaussianavatars}
\bibfield{author}{\bibinfo{person}{Shenhan Qian}, \bibinfo{person}{Tobias
  Kirschstein}, \bibinfo{person}{Liam Schoneveld}, \bibinfo{person}{Davide
  Davoli}, \bibinfo{person}{Simon Giebenhain}, {and} \bibinfo{person}{Matthias
  Nie{\ss}ner}.} \bibinfo{year}{2024}\natexlab{}.
\newblock \showarticletitle{{GaussianAvatars}: {P}hotorealistic Head Avatars
  with Rigged {3D} Gaussians}. In \bibinfo{booktitle}{\emph{Conference on
  Computer Vision and Pattern Recognition (CVPR)}}. \bibinfo{publisher}{CVF /
  IEEE}, \bibinfo{pages}{20299--20309}.
\newblock


\bibitem[Rivero et~al\mbox{.}(2024)]%
        {rivero2024rig3dgs}
\bibfield{author}{\bibinfo{person}{Alfredo Rivero}, \bibinfo{person}{ShahRukh
  Athar}, \bibinfo{person}{Zhixin Shu}, {and} \bibinfo{person}{Dimitris
  Samaras}.} \bibinfo{year}{2024}\natexlab{}.
\newblock \showarticletitle{{Rig3DGS}: {C}reating controllable portraits from
  casual monocular videos}.
\newblock \bibinfo{journal}{\emph{arXiv preprint arXiv:2402.03723}}
  (\bibinfo{year}{2024}).
\newblock


\bibitem[Rombach et~al\mbox{.}(2022)]%
        {rombach2022high}
\bibfield{author}{\bibinfo{person}{Robin Rombach}, \bibinfo{person}{Andreas
  Blattmann}, \bibinfo{person}{Dominik Lorenz}, \bibinfo{person}{Patrick
  Esser}, {and} \bibinfo{person}{Bj{\"o}rn Ommer}.}
  \bibinfo{year}{2022}\natexlab{}.
\newblock \showarticletitle{High-resolution image synthesis with latent
  diffusion models}. In \bibinfo{booktitle}{\emph{Conference on Computer Vision
  and Pattern Recognition (CVPR)}}. \bibinfo{publisher}{CVF / IEEE},
  \bibinfo{pages}{10684--10695}.
\newblock


\bibitem[Saito et~al\mbox{.}(2024a)]%
        {saito2024squeezeme}
\bibfield{author}{\bibinfo{person}{Shunsuke Saito}, \bibinfo{person}{Stanislav
  Pidhorskyi}, \bibinfo{person}{Igor Santesteban}, \bibinfo{person}{Forrest
  Iandola}, \bibinfo{person}{Divam Gupta}, \bibinfo{person}{Anuj Pahuja},
  \bibinfo{person}{Nemanja Bartolovic}, \bibinfo{person}{Frank Yu},
  \bibinfo{person}{Emanuel Garbin}, {and} \bibinfo{person}{Tomas Simon}.}
  \bibinfo{year}{2024}\natexlab{a}.
\newblock \showarticletitle{{SqueezeMe}: {E}fficient Gaussian Avatars for
  {VR}}.
\newblock \bibinfo{journal}{\emph{arXiv preprint arXiv:2412.15171}}
  (\bibinfo{year}{2024}).
\newblock


\bibitem[Saito et~al\mbox{.}(2024b)]%
        {saito2024relightable}
\bibfield{author}{\bibinfo{person}{Shunsuke Saito}, \bibinfo{person}{Gabriel
  Schwartz}, \bibinfo{person}{Tomas Simon}, \bibinfo{person}{Junxuan Li}, {and}
  \bibinfo{person}{Giljoo Nam}.} \bibinfo{year}{2024}\natexlab{b}.
\newblock \showarticletitle{Relightable gaussian codec avatars}. In
  \bibinfo{booktitle}{\emph{Conference on Computer Vision and Pattern
  Recognition (CVPR)}}. \bibinfo{publisher}{CVF / IEEE},
  \bibinfo{pages}{130--141}.
\newblock


\bibitem[Sarkar et~al\mbox{.}(2023)]%
        {sarkar2023litnerf}
\bibfield{author}{\bibinfo{person}{Kripasindhu Sarkar},
  \bibinfo{person}{Marcel~C. Buehler}, \bibinfo{person}{Gengyan Li},
  \bibinfo{person}{Daoye Wang}, \bibinfo{person}{Delio Vicini},
  \bibinfo{person}{Jérémy Riviere}, \bibinfo{person}{Yinda Zhang},
  \bibinfo{person}{Sergio Orts-Escolano}, \bibinfo{person}{Paulo Gotardo},
  \bibinfo{person}{Thabo Beeler}, {and} \bibinfo{person}{Abhimitra Meka}.}
  \bibinfo{year}{2023}\natexlab{}.
\newblock \showarticletitle{LitNeRF: Intrinsic Radiance Decomposition for
  High-Quality View Synthesis and Relighting of Faces}. In
  \bibinfo{booktitle}{\emph{ACM SIGGRAPH Asia 2023 Conference Papers, December
  12--15, 2023, Sydney, NSW, Australia}}.
\newblock
\showISBNx{979-8-4007-0315-7/23/12}
\href{https://doi.org/10.1145/3610548.3618210}{doi:\nolinkurl{10.1145/3610548.3618210}}


\bibitem[Saunders et~al\mbox{.}(2024)]%
        {saunders2024gaspgaussianavatarssynthetic}
\bibfield{author}{\bibinfo{person}{Jack Saunders}, \bibinfo{person}{Charlie
  Hewitt}, \bibinfo{person}{Yanan Jian}, \bibinfo{person}{Marek Kowalski},
  \bibinfo{person}{Tadas Baltrusaitis}, \bibinfo{person}{Yiye Chen},
  \bibinfo{person}{Darren Cosker}, \bibinfo{person}{Virginia Estellers},
  \bibinfo{person}{Nicholas Gyde}, \bibinfo{person}{Vinay~P. Namboodiri}, {and}
  \bibinfo{person}{Benjamin~E Lundell}.} \bibinfo{year}{2024}\natexlab{}.
\newblock \bibinfo{title}{{GASP}: {G}aussian Avatars with Synthetic Priors}.
\newblock
\showeprint[arxiv]{2412.07739}~[cs.CV]
\urldef\tempurl%
\url{https://arxiv.org/abs/2412.07739}
\showURL{%
\tempurl}


\bibitem[Shao et~al\mbox{.}(2024)]%
        {shao2024splattingavatar}
\bibfield{author}{\bibinfo{person}{Zhijing Shao}, \bibinfo{person}{Zhaolong
  Wang}, \bibinfo{person}{Zhuang Li}, \bibinfo{person}{Duotun Wang},
  \bibinfo{person}{Xiangru Lin}, \bibinfo{person}{Yu Zhang},
  \bibinfo{person}{Mingming Fan}, {and} \bibinfo{person}{Zeyu Wang}.}
  \bibinfo{year}{2024}\natexlab{}.
\newblock \showarticletitle{{SplattingAvatar}: {R}ealistic Real-Time Human
  Avatars with Mesh-Embedded Gaussian Splatting}. In
  \bibinfo{booktitle}{\emph{Conference on Computer Vision and Pattern
  Recognition (CVPR)}}. \bibinfo{publisher}{CVF / IEEE},
  \bibinfo{pages}{1606--1616}.
\newblock


\bibitem[Smith et~al\mbox{.}(2020)]%
        {smith2020morphable}
\bibfield{author}{\bibinfo{person}{William~AP Smith}, \bibinfo{person}{Alassane
  Seck}, \bibinfo{person}{Hannah Dee}, \bibinfo{person}{Bernard Tiddeman},
  \bibinfo{person}{Joshua~B Tenenbaum}, {and} \bibinfo{person}{Bernhard
  Egger}.} \bibinfo{year}{2020}\natexlab{}.
\newblock \showarticletitle{A morphable face albedo model}. In
  \bibinfo{booktitle}{\emph{Conference on Computer Vision and Pattern
  Recognition (CVPR)}}. \bibinfo{publisher}{CVF / IEEE},
  \bibinfo{pages}{5011--5020}.
\newblock


\bibitem[Teotia et~al\mbox{.}(2024)]%
        {teotia2024gaussianheads}
\bibfield{author}{\bibinfo{person}{Kartik Teotia}, \bibinfo{person}{Hyeongwoo
  Kim}, \bibinfo{person}{Pablo Garrido}, \bibinfo{person}{Marc Habermann},
  \bibinfo{person}{Mohamed Elgharib}, {and} \bibinfo{person}{Christian
  Theobalt}.} \bibinfo{year}{2024}\natexlab{}.
\newblock \showarticletitle{GaussianHeads: End-to-End Learning of Drivable
  Gaussian Head Avatars from Coarse-to-fine Representations}.
\newblock \bibinfo{journal}{\emph{Transactions on Graphics (TOG)}}
  \bibinfo{volume}{43}, \bibinfo{number}{6} (\bibinfo{year}{2024}),
  \bibinfo{pages}{264:1--264:12}.
\newblock


\bibitem[Wang et~al\mbox{.}(2024)]%
        {wang2024gaussianhead}
\bibfield{author}{\bibinfo{person}{Jie Wang}, \bibinfo{person}{Jiu-Cheng Xie},
  \bibinfo{person}{Xianyan Li}, \bibinfo{person}{Feng Xu},
  \bibinfo{person}{Chi-Man Pun}, {and} \bibinfo{person}{Hao Gao}.}
  \bibinfo{year}{2024}\natexlab{}.
\newblock \bibinfo{title}{GaussianHead: {H}igh-fidelity Head Avatars with
  Learnable Gaussian Diffusion}.
\newblock
\showeprint[arxiv]{2312.01632}~[cs.CV]


\bibitem[Wu et~al\mbox{.}(2024)]%
        {Wu_2024_CVPR}
\bibfield{author}{\bibinfo{person}{Guanjun Wu}, \bibinfo{person}{Taoran Yi},
  \bibinfo{person}{Jiemin Fang}, \bibinfo{person}{Lingxi Xie},
  \bibinfo{person}{Xiaopeng Zhang}, \bibinfo{person}{Wei Wei},
  \bibinfo{person}{Wenyu Liu}, \bibinfo{person}{Qi Tian}, {and}
  \bibinfo{person}{Xinggang Wang}.} \bibinfo{year}{2024}\natexlab{}.
\newblock \showarticletitle{{4D} {G}aussian Splatting for Real-Time Dynamic
  Scene Rendering}. In \bibinfo{booktitle}{\emph{Conference on Computer Vision
  and Pattern Recognition (CVPR)}}. \bibinfo{publisher}{CVF / IEEE},
  \bibinfo{pages}{20310--20320}.
\newblock


\bibitem[Wuu et~al\mbox{.}(2022)]%
        {wuu2022multiface}
\bibfield{author}{\bibinfo{person}{Cheng-hsin Wuu}, \bibinfo{person}{Ningyuan
  Zheng}, \bibinfo{person}{Scott Ardisson}, \bibinfo{person}{Rohan Bali},
  \bibinfo{person}{Danielle Belko}, \bibinfo{person}{Eric Brockmeyer},
  \bibinfo{person}{Lucas Evans}, \bibinfo{person}{Timothy Godisart},
  \bibinfo{person}{Hyowon Ha}, \bibinfo{person}{Xuhua Huang},
  \bibinfo{person}{Alexander Hypes}, \bibinfo{person}{Taylor Koska},
  \bibinfo{person}{Steven Krenn}, \bibinfo{person}{Stephen Lombardi},
  \bibinfo{person}{Xiaomin Luo}, \bibinfo{person}{Kevyn McPhail},
  \bibinfo{person}{Laura Millerschoen}, \bibinfo{person}{Michal Perdoch},
  \bibinfo{person}{Mark Pitts}, \bibinfo{person}{Alexander Richard},
  \bibinfo{person}{Jason Saragih}, \bibinfo{person}{Junko Saragih},
  \bibinfo{person}{Takaaki Shiratori}, \bibinfo{person}{Tomas Simon},
  \bibinfo{person}{Matt Stewart}, \bibinfo{person}{Autumn Trimble},
  \bibinfo{person}{Xinshuo Weng}, \bibinfo{person}{David Whitewolf},
  \bibinfo{person}{Chenglei Wu}, \bibinfo{person}{Shoou-I Yu}, {and}
  \bibinfo{person}{Yaser Sheikh}.} \bibinfo{year}{2022}\natexlab{}.
\newblock \showarticletitle{Multiface: A Dataset for Neural Face Rendering}. In
  \bibinfo{booktitle}{\emph{arXiv}}.
\newblock
\href{https://doi.org/10.48550/ARXIV.2207.11243}{doi:\nolinkurl{10.48550/ARXIV.2207.11243}}


\bibitem[Xiang et~al\mbox{.}(2024)]%
        {xiang2024flashavatar}
\bibfield{author}{\bibinfo{person}{Jun Xiang}, \bibinfo{person}{Xuan Gao},
  \bibinfo{person}{Yudong Guo}, {and} \bibinfo{person}{Juyong Zhang}.}
  \bibinfo{year}{2024}\natexlab{}.
\newblock \showarticletitle{{FlashAvatar}: {H}igh-fidelity Head Avatar with
  Efficient Gaussian Embedding}. In \bibinfo{booktitle}{\emph{Conference on
  Computer Vision and Pattern Recognition (CVPR)}}. \bibinfo{publisher}{CVF /
  IEEE}, \bibinfo{pages}{1802--1812}.
\newblock


\bibitem[Xiao et~al\mbox{.}(2024)]%
        {xiao2024bridging}
\bibfield{author}{\bibinfo{person}{Yuting Xiao}, \bibinfo{person}{Xuan Wang},
  \bibinfo{person}{Jiafei Li}, \bibinfo{person}{Hongrui Cai},
  \bibinfo{person}{Yanbo Fan}, \bibinfo{person}{Nan Xue},
  \bibinfo{person}{Minghui Yang}, \bibinfo{person}{Yujun Shen}, {and}
  \bibinfo{person}{Shenghua Gao}.} \bibinfo{year}{2024}\natexlab{}.
\newblock \showarticletitle{Bridging {3D} Gaussian and Mesh for Freeview Video
  Rendering}.
\newblock \bibinfo{journal}{\emph{arXiv preprint arXiv:2403.11453}}
  (\bibinfo{year}{2024}).
\newblock


\bibitem[Xu et~al\mbox{.}(2024)]%
        {xu2023gaussianheadavatar}
\bibfield{author}{\bibinfo{person}{Yuelang Xu}, \bibinfo{person}{Benwang Chen},
  \bibinfo{person}{Zhe Li}, \bibinfo{person}{Hongwen Zhang},
  \bibinfo{person}{Lizhen Wang}, \bibinfo{person}{Zerong Zheng}, {and}
  \bibinfo{person}{Yebin Liu}.} \bibinfo{year}{2024}\natexlab{}.
\newblock \showarticletitle{{Gaussian Head Avatar}: {U}ltra High-fidelity Head
  Avatar via Dynamic Gaussians}. In \bibinfo{booktitle}{\emph{Conference on
  Computer Vision and Pattern Recognition (CVPR)}}. \bibinfo{publisher}{CVF /
  IEEE}, \bibinfo{pages}{1931--1941}.
\newblock


\bibitem[Yang et~al\mbox{.}(2024)]%
        {yang2023deformable3dgs}
\bibfield{author}{\bibinfo{person}{Ziyi Yang}, \bibinfo{person}{Xinyu Gao},
  \bibinfo{person}{Wen Zhou}, \bibinfo{person}{Shaohui Jiao},
  \bibinfo{person}{Yuqing Zhang}, {and} \bibinfo{person}{Xiaogang Jin}.}
  \bibinfo{year}{2024}\natexlab{}.
\newblock \showarticletitle{Deformable {3D} {G}aussians for High-Fidelity
  Monocular Dynamic Scene Reconstruction}. In
  \bibinfo{booktitle}{\emph{Conference on Computer Vision and Pattern
  Recognition (CVPR)}}. \bibinfo{publisher}{CVF / IEEE},
  \bibinfo{pages}{20331--20341}.
\newblock


\bibitem[Yu et~al\mbox{.}(2024)]%
        {yu2023cogs}
\bibfield{author}{\bibinfo{person}{Heng Yu}, \bibinfo{person}{Joel Julin},
  \bibinfo{person}{Zolt{\'{a}}n~{\'{A}}d{\'{a}}m Milacski},
  \bibinfo{person}{Koichiro Niinuma}, {and}
  \bibinfo{person}{L{\'{a}}szl{\'{o}}~A. Jeni}.}
  \bibinfo{year}{2024}\natexlab{}.
\newblock \showarticletitle{{CoGS}: {C}ontrollable {G}aussian Splatting}. In
  \bibinfo{booktitle}{\emph{Conference on Computer Vision and Pattern
  Recognition (CVPR)}}. \bibinfo{publisher}{CVF / IEEE},
  \bibinfo{pages}{21624--21633}.
\newblock


\bibitem[Zhang et~al\mbox{.}(2023)]%
        {zhang2023dreamface}
\bibfield{author}{\bibinfo{person}{Longwen Zhang}, \bibinfo{person}{Qiwei Qiu},
  \bibinfo{person}{Hongyang Lin}, \bibinfo{person}{Qixuan Zhang},
  \bibinfo{person}{Cheng Shi}, \bibinfo{person}{Wei Yang}, \bibinfo{person}{Ye
  Shi}, \bibinfo{person}{Sibei Yang}, \bibinfo{person}{Lan Xu}, {and}
  \bibinfo{person}{Jingyi Yu}.} \bibinfo{year}{2023}\natexlab{}.
\newblock \showarticletitle{{DreamFace}: {P}rogressive Generation of Animatable
  {3D} Faces under Text Guidance}.
\newblock \bibinfo{journal}{\emph{Transactions on Graphics (TOG)}}
  \bibinfo{volume}{42}, \bibinfo{number}{4} (\bibinfo{year}{2023}),
  \bibinfo{pages}{138:1--138:16}.
\newblock


\bibitem[Zhang et~al\mbox{.}(2018)]%
        {8578166}
\bibfield{author}{\bibinfo{person}{Richard Zhang}, \bibinfo{person}{Phillip
  Isola}, \bibinfo{person}{Alexei~A. Efros}, \bibinfo{person}{Eli Shechtman},
  {and} \bibinfo{person}{Oliver Wang}.} \bibinfo{year}{2018}\natexlab{}.
\newblock \showarticletitle{The Unreasonable Effectiveness of Deep Features as
  a Perceptual Metric}. In \bibinfo{booktitle}{\emph{Conference on Computer
  Vision and Pattern Recognition (CVPR)}}. \bibinfo{publisher}{CVF / IEEE},
  \bibinfo{pages}{586--595}.
\newblock


\bibitem[Zhao et~al\mbox{.}(2024)]%
        {zhao2024psavatar}
\bibfield{author}{\bibinfo{person}{Zhongyuan Zhao}, \bibinfo{person}{Zhenyu
  Bao}, \bibinfo{person}{Qing Li}, \bibinfo{person}{Guoping Qiu}, {and}
  \bibinfo{person}{Kanglin Liu}.} \bibinfo{year}{2024}\natexlab{}.
\newblock \showarticletitle{{PSAvatar}: {A} point-based morphable shape model
  for real-time head avatar creation with {3D} {G}aussian splatting}.
\newblock \bibinfo{journal}{\emph{arXiv preprint arXiv:2401.12900}}
  (\bibinfo{year}{2024}).
\newblock


\bibitem[Zhuang et~al\mbox{.}(2022)]%
        {zhuang2022mofanerf}
\bibfield{author}{\bibinfo{person}{Yiyu Zhuang}, \bibinfo{person}{Hao Zhu},
  \bibinfo{person}{Xusen Sun}, {and} \bibinfo{person}{Xun Cao}.}
  \bibinfo{year}{2022}\natexlab{}.
\newblock \showarticletitle{{MoFaNeRF}: {M}orphable Facial Neural Radiance
  Field}. In \bibinfo{booktitle}{\emph{European Conference on Computer Vision
  (ECCV)}}. \bibinfo{pages}{268--285}.
\newblock


\bibitem[Zielonka et~al\mbox{.}(2024)]%
        {zielonka2024gem}
\bibfield{author}{\bibinfo{person}{Wojciech Zielonka}, \bibinfo{person}{Timo
  Bolkart}, \bibinfo{person}{Thabo Beeler}, {and} \bibinfo{person}{Justus
  Thies}.} \bibinfo{year}{2024}\natexlab{}.
\newblock \bibinfo{title}{Gaussian Eigen Models for Human Heads}.
\newblock
\showeprint[arxiv]{2407.04545}~[cs.CV]


\bibitem[Zielonka et~al\mbox{.}(2022)]%
        {Zielonka2022Insta}
\bibfield{author}{\bibinfo{person}{Wojciech Zielonka}, \bibinfo{person}{Timo
  Bolkart}, {and} \bibinfo{person}{Justus Thies}.}
  \bibinfo{year}{2022}\natexlab{}.
\newblock \showarticletitle{Instant Volumetric Head Avatars}.
\newblock \bibinfo{journal}{\emph{Conference on Computer Vision and Pattern
  Recognition (CVPR)}} (\bibinfo{year}{2022}), \bibinfo{pages}{4574--4584}.
\newblock


\bibitem[Zielonka et~al\mbox{.}(2025)]%
        {zielonka2025synthetic}
\bibfield{author}{\bibinfo{person}{Wojciech Zielonka},
  \bibinfo{person}{Stephan~J Garbin}, \bibinfo{person}{Alexandros Lattas},
  \bibinfo{person}{George Kopanas}, \bibinfo{person}{Paulo Gotardo},
  \bibinfo{person}{Thabo Beeler}, \bibinfo{person}{Justus Thies}, {and}
  \bibinfo{person}{Timo Bolkart}.} \bibinfo{year}{2025}\natexlab{}.
\newblock \showarticletitle{Synthetic Prior for Few-Shot Drivable Head Avatar
  Inversion}.
\newblock \bibinfo{journal}{\emph{arXiv preprint arXiv:2501.06903}}
  (\bibinfo{year}{2025}).
\newblock


\end{thebibliography}

\newpage

\mbox{}
\newpage

\appendix

\section{Ethics Concerns}

With any method for photorealistic face reconstruction comes the potential for misuse, especially when it comes to high resolution and fidelity methods such as ours. However, our method requires multiple high-resolution, synchronized video cameras, and capture of a variety of different expressions. As such, it is more difficult to build our model without consent of human participants, significantly reducing ethical concerns compared to (lower-quality) monocular and “single-shot” methods. However, unauthorized re-enactment of a pre-trained model is still a concern.

We also emphasize that each of the subjects of our dataset gave their informed, signed consent for the footage to be used for academic purposes.
\section{Dataset}

Here, we provide a detailed description of the datasets used for training and evaluating our method.

\subsection{Multiface}

For our training on Multiface, we use the subjects with the following IDS: 5372021, 6795937, 8870559, 002643814.

As our goal is reconstruction of the expressions, we train using frames from the E\*\*\*, GAZ\_G2 and GAZ\_G3 video sequences. 

We note that the tracked mesh provided by Multiface does not include a tongue model. As such, we *exclude* the following expressions from training:

\begin{itemize}
\item E002
\item E047-E055
\end{itemize}

Next, we select a total of 400 frames distributed across these sequences.

As our primary focus is in accurate reconstruction of expressions rather than gaze, we first select a single frame from each of the GAZ\_G2 and GAZ\_G3 video sequences if present, for a total of 34 frames. Notably, subject 5372021 does not contain any GAZ sequences.

The remaining frames are then uniformly sampled from each expression sequence from E001 to E074. Specifically, we iteratively select a single frame from each video, skip 2 frames (out of the ones provided by multiface, advancing the multiface frame index by a total of 9) and repeat until the maximum of 400 has been reached. This results in selecting a total of 8-9 frames from each video.

Finally, we manually skim through the images from a frontal camera, excluding any frames which have any amount of motion blur. This typically only includes 10-20 frames, and as such we do not replace them with additional frames.

Although Multiface also provides uv-unwrapped textures, we do not use them in our method.

Once the frames have been selected, we note that a number of cameras are pointed from the back or directly above or below the face. As such, we exclude the following camera views from training:

\begin{itemize}
    \item 400008
    \item 400010
    \item 400025
    \item 400030
    \item 400055
    \item 400067
    \item 400070
\end{itemize}

For evaluation, all frames from the EXP\_ROM7 expression capture are used.
\begin{table*}[htb]
\caption{Comparison of maximum number of Gaussians for densification.}
\centering
\begin{tabular}{ccccccccc} 
      \toprule
       & & & \multicolumn{3}{c}{Test Reenactment} & \multicolumn{3}{c}{Novel View}\\
      \hline
       & FPS & Landmark $\downarrow$ & LPIPS $\downarrow$ & SSIM $\uparrow$ & PSNR $\uparrow$ & LPIPS $\downarrow$ & SSIM $\uparrow$ & PSNR $\uparrow$ \\
      \hline
      200K & 28.0 & 49.0 $\pm$ 49.2 & 0.207 $\pm$ 0.076 & 0.770 $\pm$ 0.055 & 24.3 $\pm$ 1.4 & 0.158 $\pm$ 0.062 & 0.831 $\pm$ 0.042 & 26.1 $\pm$ 1.5 \\
      500K & 15.3 & 34.7 $\pm$ 33.6 & 0.174 $\pm$ 0.068 & 0.776 $\pm$ 0.057 & 24.5 $\pm$ 1.3 & 0.135 $\pm$ 0.058 & 0.842 $\pm$ 0.058 & 26.6 $\pm$ 1.6\\
      1M & 7.8 & 33.8 $\pm$ 30.2 & 0.161 $\pm$ 0.062 & 0.775 $\pm$ 0.060 & 24.4 $\pm$ 1.3 & 0.127 $\pm$ 0.054 & 0.843 $\pm$ 0.048 & 26.5 $\pm$ 1.6 \\
      2M & 4.2 & 30.7 $\pm$ 29.6 & 0.152 $\pm$ 0.057 & 0.776 $\pm$ 0.061 & 24.4 $\pm$ 1.3 & 0.122 $\pm$ 0.051 & 0.848 $\pm$ 0.042 & 26.7 $\pm$ 1.7 \\
      4M & 2.4 & 30.5 $\pm$ 28.9 & 0.150 $\pm$ 0.056 & 0.780 $\pm$ 0.062 & 24.4 $\pm$ 1.3 & 0.122 $\pm$ 0.048 & 0.846 $\pm$ 0.042 & 26.6 $\pm$ 1.6\\
      6M & 2.2 & 31.7 $\pm$ 28.6 & 0.150 $\pm$ 0.055 & 0.779 $\pm$ 0.061 & 24.4 $\pm$ 1.2 & 0.123 $\pm$ 0.045  & 0.848 $\pm$ 0.038 &  26.7 $\pm$ 1.5 \\ 

      \bottomrule
\end{tabular}%
\label{tab:performance}
\end{table*}

\begin{figure*}[htb]
    \centering
    \includegraphics[width=0.91\textwidth]{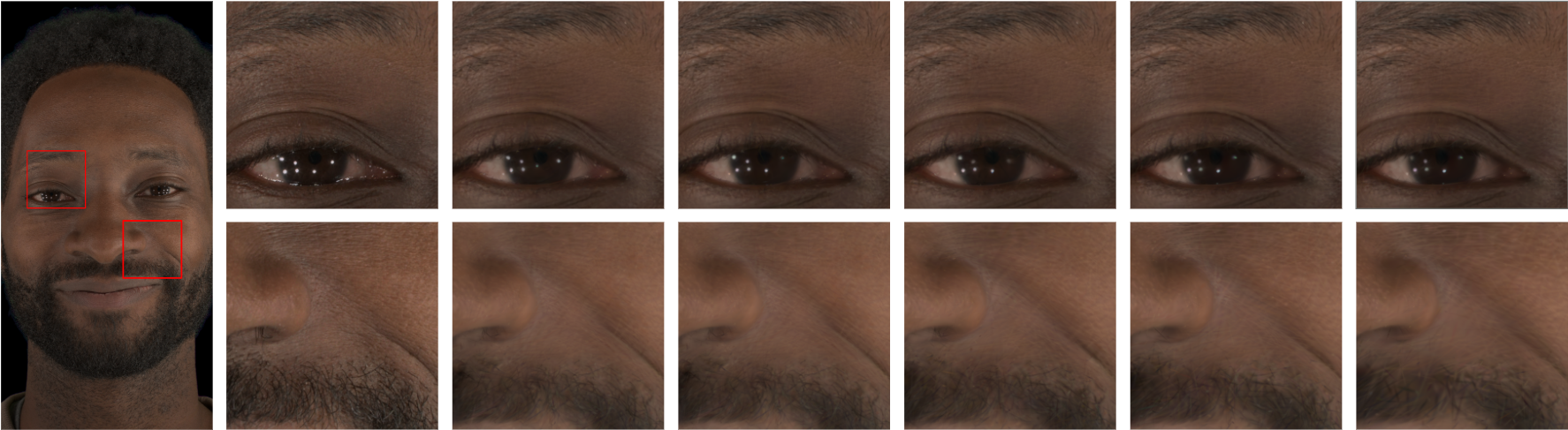}
    
    \vspace{-1.0mm}
    
    \makebox[0.12\textwidth][c]{}\hspace{0.01\textwidth}\makebox[0.12\textwidth][c]{GT}\hspace{0.01\textwidth}\makebox[0.12\textwidth][c]{6M}\hspace{0.01\textwidth}\makebox[0.12\textwidth][c]{2M}\hspace{0.01\textwidth}\makebox[0.12\textwidth][c]{1M}\hspace{0.01\textwidth}\makebox[0.12\textwidth][c]{500K}\hspace{0.01\textwidth}\makebox[0.12\textwidth][c]{200K}
    
    \vspace{-2.0mm}
    
    \caption{Comparison between maximum number of Gaussians for densification.}
    \label{fig:GS_count}
\end{figure*}
\begin{figure}
    \centering
    \includegraphics[width=0.9\columnwidth]{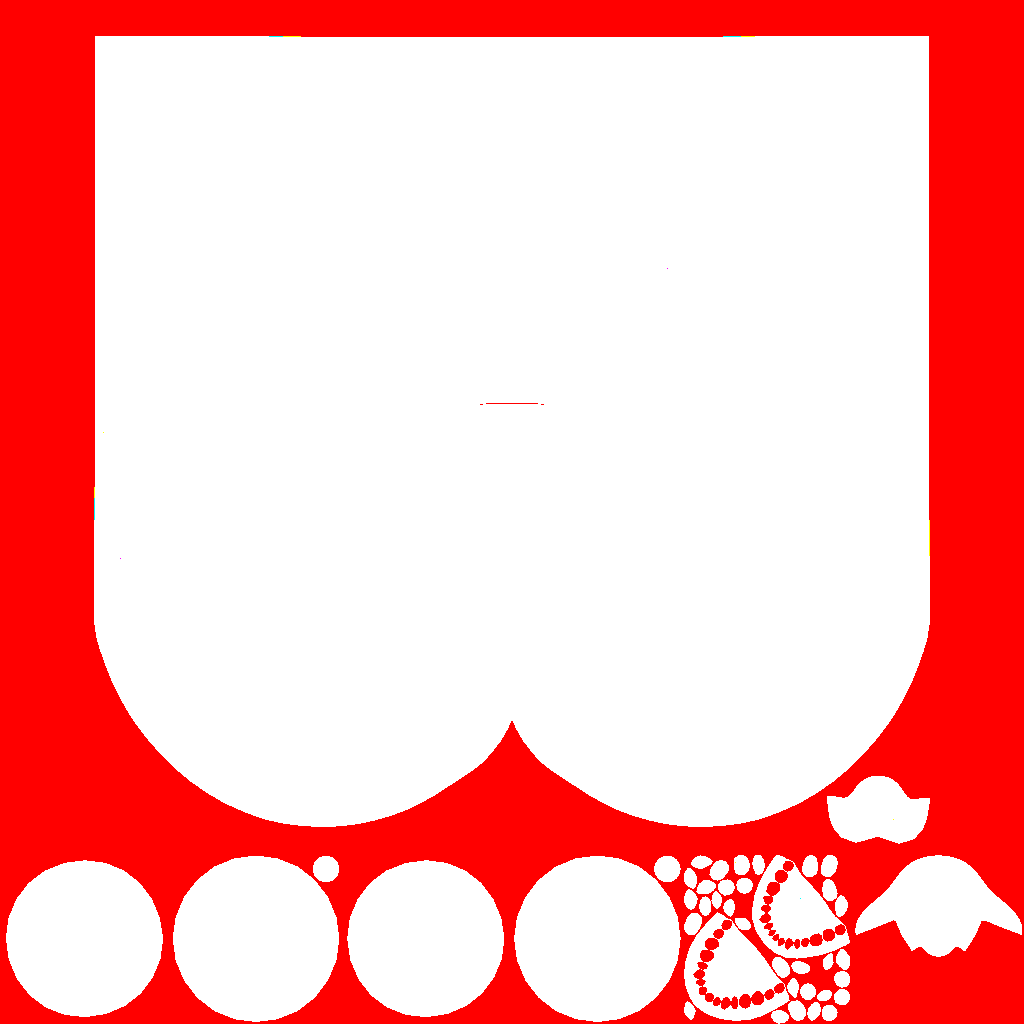}
    \caption{Our UV layout: red areas indicate invalid UV locations that are not covered by any triangle.}
    \label{fig:uv_layout}
\end{figure}

\section{UV Layout}

In Figure \ref{fig:uv_layout}, we show the UV Layout of our 3DMM. All invalid areas which are not occupied by any triangles are marked in red. Any Gaussians initialized in these areas are projected to the closest triangle.

The main facial area consists of the large plane above. The four larger circles and 2 small circles on the lower left are used to represent the two eyes. The cluster of shapes in the lower center right represent the teeth, and the remaining two shapes on the lower right represent the mouth interior.

\section{Performance Analysis}

\subsection{Hardware}

Our models were all trained using a single Nvidia H100, with 80 GB of GPU memory.

\subsection{Comparison of Gaussian count}

We additionally provide a comparison of the runtime performance of our model with an upper bound of Gaussians. Specifically, we adjust the densification algorithm to prevent further densification when the upper bound of Gaussians would be exceeded. The frame rate is roughly inversely proportional to the number of Gaussians, indicating that each individual Gaussian incurs a similar amount of computational cost. 

As can be seen in Table \ref{tab:performance} and Figure \ref{fig:GS_count}, image quality improves up to 2M total Gaussians, after which the quality stagnates.
Specifically, while overall quality remains high even down to 200K, some finer details, degrade in quality as the number of Gaussians are insufficient in that area.

Although we set an upper limit of at most 6M Gaussians, for most subjects the total count does not exceed 4M, which explains why there no significant difference in terms of runtime performance between the two.

\section{Network Architecture}

In this section, we provide the details of our network architecture.

\subsection{U-Net}

\begin{figure}
    \centering
    \includegraphics[width=0.9\columnwidth]{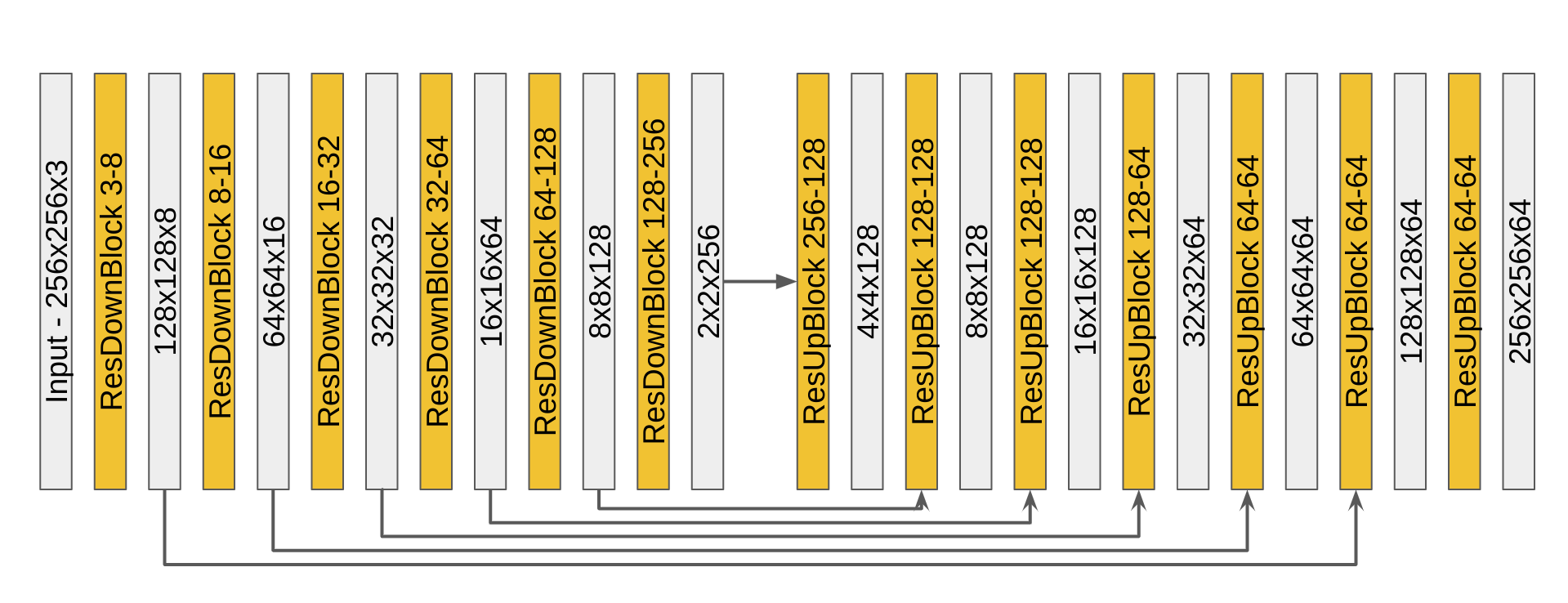}
    \caption{A schematic of our U-Net architecture}
    \label{fig:Unet}
\end{figure}

\begin{figure}
    \centering
    \includegraphics[width=0.9\columnwidth]{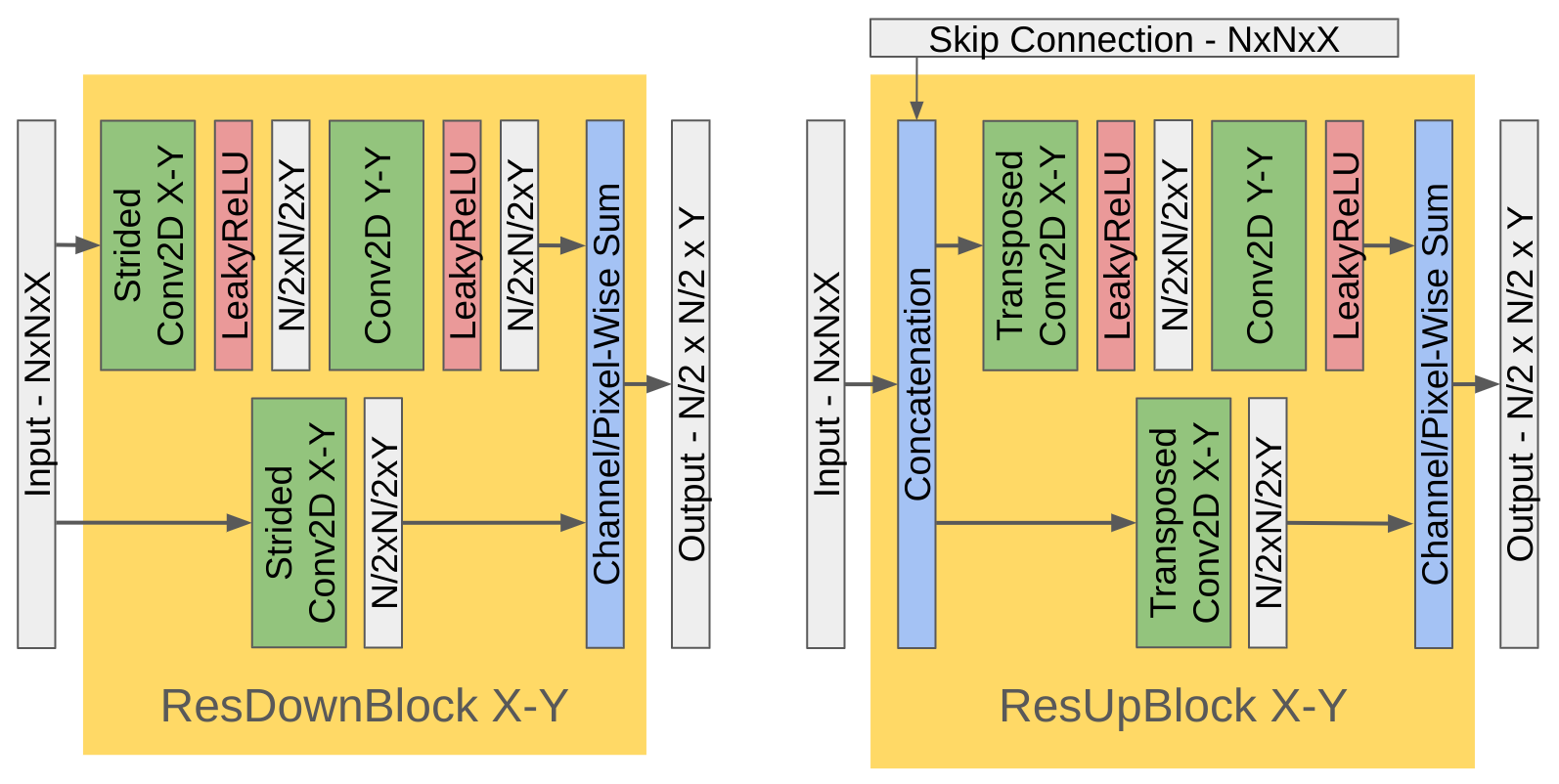}
    \caption{A schematic of the blocks of our UNet}
    \label{fig:UnetBlock}
\end{figure}

For our U-Net, we use 6 downsampling and 6 upsampling blocks, with skip connections between equal resolution layers as illustrated in Figure \ref{fig:Unet}.

Each block in the Unet consists of two convolutional layers, the first layer downsampling or upsampling using striding or transposing respectively. These two layers are each followed by a Leaky ReLU with alpha value 0.2, and a skip connection using a strided or transposed convolutional layer with no activation, as illustrated in Figure \ref{fig:UnetBlock}. All layer weights are uniform initialized using the default pyTorch settings.

\subsection{MLPs}

\begin{figure}
    \centering
    \includegraphics[width=0.9\columnwidth]{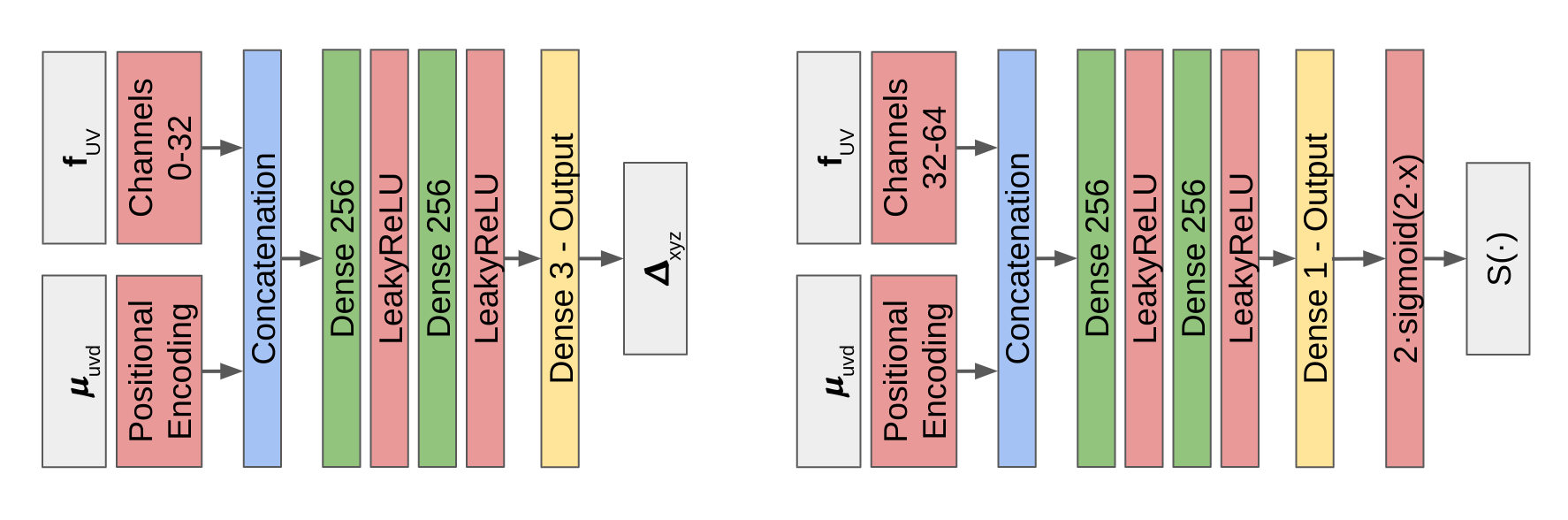}
    \caption{A schematic of our MLP heads}
    \label{fig:MLP}
\end{figure}

For our deformation and shading MLP heads, we use 2 fully connected  256-wide hidden layers, and a single fully connected layer with output size of 3 and 1 respectively. The first 32 channels of the U-Net output are used for the deformation network input, while the last 32 channels are used for the shading network input, avoiding the smoothness regularizer applied onto the deformation from excessively affecting the shading network. 

The positions of each Gaussian in UVD space is used as a secondary input, and is positionally encoded using 8 frequencies.

While the output of the deformation network is used directly, the output of the shading network has the following function applied to it:
\begin{equation}
 f(x) = 2 \frac{1}{1+e^{2x}}
\end{equation}
with the network output first being multiplied by two, which we empirically found to improve the quality and intensity of the shading. We then apply the sigmoid function, which has a range of [0, 1]. Finally, we multiply the result by 2 again, ensuring that the final output consists of the range [0, 2]. This centers the function around the value 1, which we assume is the dominate value, avoiding potential issues with vanishing gradients.

All layers except the output layers are initialized using the default pyTorch settings. The final output layers, labeled as such and marked in yellow, have their weights initialized as zero. This guarantees that with the initialized weights, the deformation network will always output zero deformation and the shading network will always output a shading value of 1. As we only begin training and using the deformation and shading networks after 10K iterations of first pre-training the Gaussians, this avoids a potential sudden change in color and shading, as the networks are guaranteed to have no effect when run with the initial weights, allowing for smoother training.

\subsection{CNN/MLP Ablation}

For one of our ablations, we used an MLP rather than U-Net model. To do so, we added two additional downsampling blocks, and removed all upsampling blocks, replacing the U-Net with a CNN encoder which produces a single global latent code. The deformation MLP was then extended from 2 hidden layers, to 8 hidden layers with a skip connection between the input and the fourth hidden layer.

However, the eight-layer MLP required significantly more memory than our UNet/MLP hybrid. As such, we upper bound the total number of Gaussians to 1.5 million, and use bfloat16 mixed precision for the 8 hidden layers. Without both these changes, the 80 GB of memory on an H100 would not be sufficient for training whole images at a time.
\section{Jacobian Covariance transformation}

We also ran an ablation test of our model in which we replaced our Jacobian-based covariance transformation (computed via automatic differentiation) with an explicit scaled rigid transformation for each triangle. This variant of our method applies a transformation to the canonical UVD Gaussians that is therefore similar to that in GaussianAvatars. In adapting the GaussianAvatar rigging transformation to our UVD canonical representation, we keep their rotation formulation (basing it off of the orientation of each triangle) and we define the scale factor as the ratio of the size of world-space and UV-space triangles. As this new Jacobian does not account for our additional (residual) deformation field, we disable this field for a fair comparison. Note that, although we implement the transform suggested by GaussianAvatars, here we compare two versions of our method without the residual field and with canonical Gaussians that still move within a continuous UVD space and, thus, also move across mesh triangles. 

\begin{figure}[htb]
    \centering
    \includegraphics[width=0.99\columnwidth]{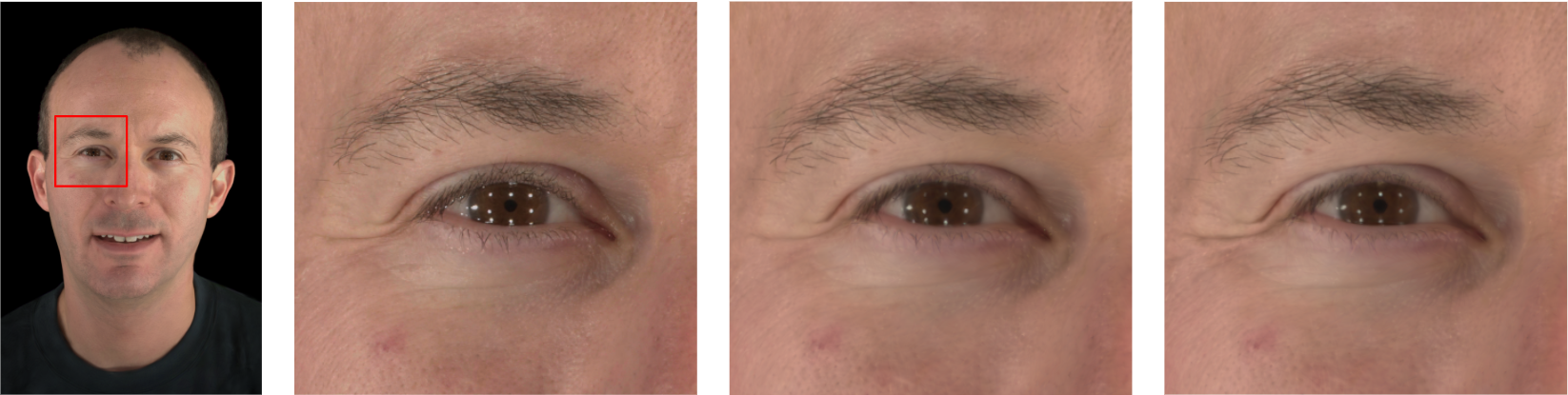}
    
    \vspace{-1.0mm}

\makebox[0.166\columnwidth][c]{GT}\hspace{0.02\columnwidth}\makebox[0.254\columnwidth][c]{GT}\hspace{0.02\columnwidth}\makebox[0.254\columnwidth][c]{Scaled Rigid}\hspace{0.02\columnwidth}\makebox[0.254\columnwidth][c]{Ours (No Warp)}

    \vspace{-2.0mm}
    
    \caption{A comparison of our Jacobian transformation with a scaled rigid transformation}
    \label{fig:jac_comp}
\end{figure}

The new baseline with the alternative, analytic Jacobian presents similar level of quality compared to our method, as can be seen in Table \ref{tab:jaccompare} and Figure \ref{fig:jac_comp}. We believe that this is due to the fact that in practice, on the scale of individual Gaussians, the nonrigid component of the face transformation is sufficiently small and this locally-rigid approximation is still accurate enough in order to produce comparable details.

\begin{table}[htb]
\caption{Comparison between our Jacobian-based and Scaled Rigid transformation}
\resizebox{\columnwidth}{!}{%
\centering

\begin{tabular}{ccccc}
      \toprule
      & & \multicolumn{3}{c}{Test Reenactment}\\
      \hline
        & Landmark $\downarrow$ & LPIPS $\downarrow$ & SSIM $\uparrow$ & PSNR $\uparrow$\\
      \hline
      Scaled Rigid & 30.3 $\pm$ 11.6 & 0.146 $\pm$ 0.014 & 0.786 $\pm$ 0.027 & 25.1 $\pm$ 1.2 \\
      Ours (No Warp) & 30.1 $\pm$ 12.2 & 0.145 $\pm$ 0.015 & 0.786 $\pm$ 0.027 & 25.0 $\pm$ 1.4 \\
      \bottomrule
\end{tabular}%
}
\label{tab:jaccompare}
\end{table}

However, we emphasize that our approach is much easier to implement and extend with additional components such as the residual deformation field. We further note that, in the comparison, the positions of each Gaussian are still transformed using our nonrigid UVD-based mapping, and we still maintain the ability to move Gaussians across triangle boundaries, both of which significantly improve the overall effectiveness of our method over GaussianAvatars.



\section{Dataset split results}

We also provide separate quantitative evaluations for each dataset, in order to show that our method outperforms all other settings for both datasets individually, as well as across the board. The resulting metrics are given in Table \ref{tab:test1} and \ref{tab:test2}.

\begin{table*}[htb]
\caption{Quantitative comparison on our dataset; metrics evaluated on the full head. \hlredcaption{Best}, \hlorangecaption{second best}, \hlyellowcaption{third best} scores are highlighted.}
\centering
\begin{tabular}{cccccccc} 
      \toprule
      & & \multicolumn{3}{c}{Test Reenactment} & \multicolumn{3}{c}{Novel View} \\
      \hline
        & Landmark $\downarrow$ & LPIPS $\downarrow$ & SSIM $\uparrow$ & PSNR $\uparrow$ & LPIPS $\downarrow$ & SSIM $\uparrow$ & PSNR $\uparrow$ \\
      \hline
      No Warp & 30.1 $\pm$ 12.2 & 0.145 $\pm$ 0.014 & \hlyellow{0.786 $\pm$ 0.027} & \hlorange{25.0 $\pm$ 1.4} & 0.126 $\pm$ 0.011 & 0.811 $\pm$ 0.019 & 26.2 $\pm$ 1.9\\ 
      No Shading & 30.2 $\pm$ 12.4 & 0.127 $\pm$ 0.015 & 0.771 $\pm$ 0.032 & 24.2 $\pm$ 1.4 & 0.105 $\pm$ 0.010 & 0.833 $\pm$ 0.020 & 26.2 $\pm$ 1.5 \\ 
      No Network & 35.3 $\pm$ 13.4 & 0.160 $\pm$ 0.016 & 0.768 $\pm$ 0.028 & 23.7 $\pm$ 1.6 & 0.136 $\pm$ 0.013 & 0.795 $\pm$ 0.021 & 24.7 $\pm$ 2.0\\ 
      No Densify & 68.6 $\pm$ 19.3 & 0.223 $\pm$ 0.020 & 0.780 $\pm$ 0.028 & 24.4 $\pm$ 1.2 & 0.192 $\pm$ 0.012 & 0.803 $\pm$ 0.016 & 25.9 $\pm$ 1.1 \\
      MLP & \hlyellow{29.0 $\pm$ 11.5} & 0.128 $\pm$ 0.015 & 0.781 $\pm$ 0.030 & 24.8 $\pm$ 1.3 & 0.104 $\pm$ 0.011 & 0.832 $\pm$ 0.026 & 26.7 $\pm$ 1.3 \\
      No VGG & 30.3 $\pm$ 11.1 & 0.145 $\pm$ 0.017 & \hlred{0.787 $\pm$ 0.034} & \hlred{25.0 $\pm$ 1.4} & 0.118 $\pm$ 0.010 & \hlorange{0.850} $\pm$ 0.021 & \hlyellow{27.1} $\pm$ 1.3 \\
      No Triangle Updates & 30.6 $\pm$ 11.4 & \hlred{0.119 $\pm$ 0.015} & 0.779 $\pm$ 0.033 & 24.8 $\pm$ 1.3 & \hlred{0.096 $\pm$ 0.010} & \hlred{0.854} $\pm$ 0.023 & \hlred{27.3 $\pm$ 1.3} \\
      \hline
      GaussianAvatars & 65.3 $\pm$ 25.9 & 0.217 $\pm$ 0.022 & 0.783 $\pm$ 0.028 & 24.6 $\pm$ 1.6 & 0.157 $\pm$ 0.009 & 0.833 $\pm$ 0.014 & 25.7 $\pm$ 1.9 \\
      RGCA  & 140.5 $\pm$ 81.9  & 0.182 $\pm$ 0.020 & 0.753 $\pm$ 0.031 & 23.3 $\pm$ 1.2 & 0.129 $\pm$ 0.009 & 0.845 $\pm$ 0.012 & 26.2 $\pm$ 1.9 \\
      GaussianHeadAvatar & 1655.1 $\pm$ 628.5 & 0.198 $\pm$ 0.034 & 0.694 $\pm$ 0.046 & 14.5 $\pm$ 2.1 & 0.151 $\pm$ 0.011 & 0.727 $\pm$ 0.017 & 15.8 $\pm$ 2.1 \\
      MVP & 229.0 $\pm$ 173.4 &0.240 $\pm$ 0.038 & 0.742 $\pm$ 0.043 & 23.2 $\pm$ 1.6 & 0.202 $\pm$ 0.013 & 0.794 $\pm$ 0.014 & 25.9 $\pm$ 1.3 \\
      \hline
      Ours  & \hlred{25.7 $\pm$ 10.5} & \hlorange{0.120 $\pm$ 0.015} & 0.781 $\pm$ 0.033 & \hlyellow{24.9 $\pm$ 1.4} & \hlorange{0.099 $\pm$ 0.010}  & \hlyellow{0.846 $\pm$ 0.023} &  \hlorange{27.1 $\pm$ 1.3} \\ 
      Ours (1M GS.) & \hlorange{26.6} $\pm$ 10.0 & \hlyellow{0.125 $\pm$ 0.015} & 0.781 $\pm$ 0.033 & 24.7 $\pm$ 1.3 & \hlyellow{0.101 $\pm$ 0.011}  & 0.843 $\pm$ 0.024 & 27.0 $\pm$ 1.3 \\ 
      Ours (200K GS.) & 37.2 $\pm$ 19.4 & 0.162 $\pm$ 0.015 & \hlorange{0.783} $\pm$ 0.028 & 24.8 $\pm$ 1.3 & 0.133 $\pm$ 0.011  & 0.828 $\pm$ 0.020 & 26.5 $\pm$ 1.1 \\
      \bottomrule
\end{tabular}%
\label{tab:test1}
\end{table*}

\begin{table*}[htb]
\caption{Quantitative comparison on Multiface \cite{wuu2022multiface}; metrics evaluated on the full head. \hlredcaption{Best}, \hlorangecaption{second best}, \hlyellowcaption{third best} scores are highlighted.}
\centering
\begin{tabular}{cccccccc} 
      \toprule
      & & \multicolumn{3}{c}{Test Reenactment} & \multicolumn{3}{c}{Novel View} \\
      \hline
        & Landmark $\downarrow$ & LPIPS $\downarrow$ & SSIM $\uparrow$ & PSNR $\uparrow$ & LPIPS $\downarrow$ & SSIM $\uparrow$ & PSNR $\uparrow$ \\
      \hline
      No Warp & 46.0 $\pm$ 46.2 & \hlorange{0.208 $\pm$ 0.025} & \hlyellow{0.780 $\pm$ 0.028} & \hlyellow{23.7 $\pm$ 0.4} & \hlorange{0.186 $\pm$ 0.013} & 0.816 $\pm$ 0.030 & 25.8 $\pm$ 0.7 \\ 
      No Shading & \hlyellow{44.2 $\pm$ 45.9} & \hlyellow{0.210 $\pm$ 0.027} & 0.776 $\pm$ 0.030 & 23.5 $\pm$ 0.5 & \hlyellow{0.188 $\pm$ 0.012} & \hlorange{0.835 $\pm$ 0.025} & 25.7 $\pm$ 0.7 \\ 
      No Network & 51.4 $\pm$ 46.4 & 0.219 $\pm$ 0.027 & 0.770 $\pm$ 0.030 & 23.4 $\pm$ 0.5 & 0.197 $\pm$ 0.015 & 0.807 $\pm$ 0.035 & 25.1 $\pm$ 0.6 \\ 
      No Densify & 78.3 $\pm$ 53.7 & 0.346 $\pm$ 0.032 & 0.721 $\pm$ 0.024 & 22.3 $\pm$ 0.4 & 0.286 $\pm$ 0.009 & 0.766 $\pm$ 0.022 & 23.9 $\pm$ 0.5 \\
      MLP  & 47.4 $\pm$ 46.1 & 0.213 $\pm$ 0.027 & 0.778 $\pm$ 0.030 & 23.5 $\pm$ 0.5 & 0.189 $\pm$ 0.015 & 0.811 $\pm$ 0.030 & 25.3 $\pm$ 0.6 \\
      No VGG & 46.6 $\pm 48.7$ & 0.230 $\pm$ 0.024 & \hlred{0.789$\pm$ 0.030} & \hlred{23.9 $\pm$ 0.5} & 0.209 $\pm$ 0.011 & \hlred{0.839 $\pm$ 0.023} & \hlorange{26.1 $\pm$ 0.7} \\
      No Triangle Updates & \hlred{38.3 $\pm$ 43.2} & 0.212 $\pm$ 0.029 & 0.778 $\pm$ 0.031 & 23.7 $\pm$ 0.5 & 0.192 $\pm$ 0.012 & 0.831 $\pm$ 0.023 & \hlred{26.1 $\pm$ 0.7} \\
      \hline
      GaussianAvatars  & 88.1 $\pm$ 62.7 & 0.313 $\pm$ 0.026 & 0.754 $\pm$ 0.025 & 23.2 $\pm$ 0.4 & 0.275 $\pm$ 0.010 & 0.783 $\pm$ 0.026 & 24.9 $\pm$ 0.9 \\
      RGCA  & 65.4 $\pm$ 56.5 & 0.271 $\pm$ 0.028 & 0.684 $\pm$ 0.035 & 22.3 $\pm$ 0.7 & 0.229 $\pm$ 0.009 & 0.827 $\pm$ 0.022 & 25.6 $\pm$ 0.9 \\
      GaussianHeadAvatar & 381.5 $\pm$ 201.6 & 0.318 $\pm$ 0.039 & 0.616 $\pm$ 0.027 & 18.3 $\pm$ 0.6 & 0.291 $\pm$ 0.011 & 0.629 $\pm$ 0.013 & 18.7 $\pm$ 0.8 \\
      MVP & 131.1 $\pm$ 72.0 & 0.350 $\pm$ 0.030 & 0.704 $\pm$ 0.022 & 23.3 $\pm$ 0.7 & 0.333 $\pm$ 0.006 & 0.747 $\pm$ 0.009 & 25.6 $\pm$ 1.0 \\
      \hline
      Ours & \hlorange{39.9 $\pm$ 46.2} & \hlred{0.202 $\pm$ 0.027} & \hlorange{0.782 $\pm$ 0.030} & \hlorange{23.8 $\pm$ 0.5} & \hlred{0.178 $\pm$ 0.013} & \hlyellow{0.834 $\pm$ 0.025} & \hlyellow{26.0 $\pm$ 0.6} \\
      Ours (1M GS.) & 48.3 $\pm$ 42.9 & 0.233 $\pm$ 0.030 & 0.764 $\pm$ 0.030 & 23.4 $\pm$ 0.6 & 0.197 $\pm$ 0.014 & 0.810 $\pm$ 0.033 &  25.4 $\pm$ 0.3 \\
      Ours (200K GS.) & 72.8 $\pm$ 51.2 & 0.296 $\pm$ 0.037 & 0.743 $\pm$ 0.031 & 23.1 $\pm$ 0.4 & 0.340 $\pm$ 0.012 & 0.797 $\pm$ 0.026 &  25.0 $\pm$ 1.0 \\ 

      \bottomrule
\end{tabular}%
\label{tab:test2}
\end{table*}

\end{document}